\documentclass[11pt,a4paper,onecolumn,copyright]{handshaketemplate}
\usepackage[authoryear,round]{natbib}

\usepackage{hyperref}
\usepackage{url}

\usepackage{graphicx} % Required for inserting images
\usepackage[utf8]{inputenc} % allow utf-8 input
\usepackage[T1]{fontenc}    % use 8-bit T1 fonts
\usepackage{hyperref}       % hyperlinks
\usepackage{url}            % simple URL typesetting
\usepackage{booktabs}       % professional-quality tables
\usepackage{amsfonts}       % blackboard math symbols
\usepackage{nicefrac}       % compact symbols for 1/2, etc.
\usepackage{microtype}      % microtypography
\usepackage{xcolor}         % colors
\usepackage[normalem]{ulem} % \sout strikethrough (used by \critdel)
\usepackage{multirow}
\usepackage{pgfplots}
\usepackage{tikz}
\usepackage{float}
\usepackage{needspace}

\usetikzlibrary{calc}
\pgfplotsset{compat=1.18}
\usepackage[skins,breakable]{tcolorbox}
\newtcolorbox{promptbox}[1]{
  breakable, colback=gray!4, colframe=gray!55!black,
  fonttitle=\bfseries, title={#1},
  boxrule=0.4pt, left=4pt, right=4pt, top=3pt, bottom=3pt}

\newtcolorbox{evolutionbox}[1]{
  enhanced, breakable,
  colback=green!3,
  colframe=green!35!black,
  colbacktitle=green!35!black,
  coltitle=white,
  fonttitle=\bfseries,
  fontupper=\small,
  title={#1},
  boxrule=0.45pt,
  arc=1pt,
  left=6pt, right=6pt,
  top=5pt, bottom=5pt}

\newtcolorbox{evolutionpagebox}[1]{
  enhanced, colback=green!3, colframe=green!35!black,
  colbacktitle=green!35!black, coltitle=white,
  fonttitle=\bfseries, fontupper=\small,
  title={#1}, boxrule=0.45pt, arc=1pt,
  left=8pt, right=8pt, top=7pt, bottom=7pt}

\usepackage{xcolor}
\usepackage{enumitem}
\usepackage[normalem]{ulem}
\newcommand{\comparisonhl}[1]{\begingroup\setlength{\fboxsep}{1pt}\colorbox{yellow!28}{#1}\endgroup}
\newtcolorbox{responsepanel}[2]{
  enhanced,equal height group=healthbenchresponses,
  colback=white,colframe=gray!50,colbacktitle=gray!10,coltitle=black,
  title={\textbf{#1}\\{\footnotesize Physician-authored rubric score: \textbf{#2}}},
  fonttitle=\fontsize{9}{10.5}\selectfont,
  fontupper=\fontsize{8}{9.3}\selectfont,
  before upper={\raggedright\setlength{\parskip}{3pt}},
  boxrule=0.4pt,arc=1pt,left=5pt,right=5pt,top=5pt,bottom=5pt,
  before skip=0pt,after skip=0pt}
\usepackage{listings}

\lstdefinestyle{promptstyle}{
  basicstyle=\footnotesize\ttfamily,
  breaklines=true,
  breakatwhitespace=true,
  breakautoindent=false,
  breakindent=0pt,
  columns=fullflexible,
  keepspaces=true,
  showstringspaces=false,
  showspaces=false,
  showtabs=false,
  upquote=true,
  tabsize=2,
  aboveskip=2pt, belowskip=2pt,
  escapeinside={(*@}{@*)},
  literate={—}{{\normalfont\textemdash}}1 {–}{{\normalfont\textendash}}1 {`}{{\char96}}1,
}

\newcommand{\cornerlogo}[2][1]{%
  \begin{tikzpicture}[remember picture,overlay]
    \node[anchor=north west, xshift=2.0cm, yshift=-1.8cm]
      at (current page.north west)
      {\includegraphics[scale=#1]{#2}};
    \draw[line width=0.3pt]
      ($(current page.north west)+(2.0cm,-2.7cm)$) --
      ($(current page.north east)+(-2.0cm,-2.7cm)$);
  \end{tikzpicture}%
}

\newcommand{\promptfile}[2]{%
  \begin{promptbox}{#1}%
    \lstinputlisting[style=promptstyle]{#2}%
  \end{promptbox}}

\definecolor{critgreen}{rgb}{0.0,0.5,0.0}
\definecolor{critred}{rgb}{0.7,0.0,0.0}
\newcommand{\critadd}[1]{\item[\textcolor{black}{\textbf{+}}] \textcolor{black}{#1}}
\newcommand{\critdel}[1]{\item[\textcolor{black}{\textbf{--}}] \textcolor{black}{\sout{#1}}}

\makeatletter
\renewcommand\AB@authnote[1]{}
\renewcommand\AB@affilnote[1]{}
\makeatother

\newcommand{\rubricon}{Rubricon}

\title{Rubricon: Automated Rubric Generation and Refinement for LLM Evaluation and Training}

\author{Ulyana Tkachenko\textsuperscript{*}}
\author{Hui Wen Goh\textsuperscript{*}}
\author{Guram Gogia}
\author{Jonas Mueller}

\affil{Handshake AI}
\begin{abstract}
Rubrics enable reward assignment on tasks without verifiable answers, but are expensive to manually produce from scratch. We present \emph{\rubricon{}}, a pipeline that generates discriminative rubrics without requiring human preference labels, rubric annotation, or generator-model training. Using any LLM, \rubricon{} samples multiple rubrics, selects the most discriminative candidate, adds criteria to capture overlooked requirements or response flaws, and prunes criteria that are misaligned or shallow. Alongside LLM self-critique and test-time scaling, \rubricon{} uses observed criterion pass rates on example model responses to guide refinement. Across five professional, consumer, clinical, and general-use LLM benchmarks, \rubricon{}  produces rubrics whose scores of frontier LLM responses exhibit \textbf{9\%} greater alignment with labor-intensive expert-authored rubrics, compared to the best alternative synthetic rubric generation method among RRD, QSRG, OpenRubrics, and agentic generation via harnesses like Codex. On HealthBench, post-training a Qwen3.5-9B model with \rubricon{} rubrics yields \textbf{4.5\%} better performance than training with rubrics from the strongest alternative synthetic rubric generator, RRD.

\end{abstract}

\begin{document}
\cornerlogo[0.15]{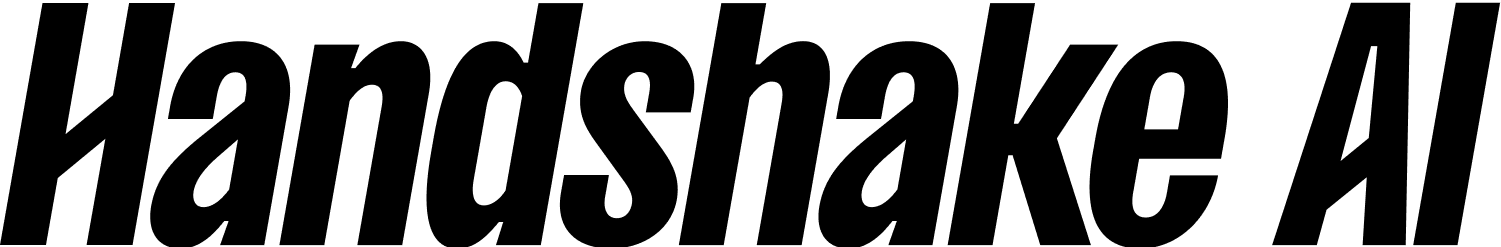} % removed per camera-ready formatting

\maketitle

\begingroup
\renewcommand{\thefootnote}{\fnsymbol{footnote}}
\footnotetext[1]{Equal contribution}
\endgroup

\begingroup \makeatletter \renewcommand\@makefnmark{} 
\footnotetext{\hspace*{-0.5em} Correspondence to \texttt{\{ulyana.tkachenko,huiwen.goh,jonas.mueller\}@joinhandshake.com}} 
\makeatother \endgroup

\section{Introduction}

Rubrics provide explicit, checkable criteria for systematically evaluating open-domain LLM responses and assigning post-training rewards~\citep{gunjal2025rubrics,
huang2025reinforcement, xu2026alternating}. Expert-authored rubrics are the gold standard but are
painstaking to produce (datasets with rubrics can require hundreds of
physicians or bankers~\citep{arora2025healthbench, bankertoolbench}). Asking an LLM to generate a rubric yields criteria too shallow to separate strong responses from weak ones, or even criteria that incorrectly favor lower-quality responses~\citep{shen2026rethinking, liu2025openrubrics}. 
A further challenge is building rubrics that remain discriminative as models improve, which requires criteria that will appropriately penalize flaws in responses from stronger \emph{future} models not yet available.

We propose \emph{\rubricon{}}, a straightforward pipeline that automatically generates higher-quality rubrics via three
stages: \textbf{Generate-and-Select}, \textbf{Add}, and \textbf{Prune}. It targets three properties a
good rubric needs: \textbf{(i)} the ability to discriminate between responses of different quality, \textbf{(ii)} coverage so that substantively weak responses do not go unpenalized, and
\textbf{(iii)} headroom for quantifying flaws in responses from stronger future models. The pipeline does not require any human data annotation or model fine-tuning, and can utilize \emph{any} LLM to generate/refine the rubrics, yielding better rubrics as LLMs advance.
% Instead, its primary signal is the \emph{strong--weak edge}: the difference in criterion pass rates between stronger- and weaker-model responses.
\rubricon{} also supports iterative refinement as the Add and Prune stages can be repeatedly applied to continuously reassess and improve a rubric's criteria.

\rubricon{} illustrates a way to scale \emph{test-time compute} \citep{snell2024scaling} for rubric generation that is empirically more effective than alternate approaches like agentic rubric generation or Recursive Rubric Decomposition \citep{shen2026rethinking}. Relying solely on LLM self-critique or ensembling yields limited test-time scaling gains due to LLM brittleness and biases in the rubric generation setting. Instead, \rubricon{} additionally grounds its refinement stages in how candidate rubrics actually score example responses. Because existing LLMs cannot properly self-judge their own generated rubrics, our refinement stages encourage the models to rely on particular outside signals and workflows which simplify the judgment calls a model must make. Compared to existing rubric-generation techniques, the resulting \rubricon{}-generated rubrics score responses more consistently with expert-authored rubrics (Table \ref{tab:summary_main}) and improve post-training outcomes (Figure \ref{fig:healthbench-rubric-source}).

\section{Related Work}
\label{rel_works}

Synthetic rubrics offer a structured alternative to holistic evaluation, providing explicit, interpretable criteria for open-ended tasks and post-training rewards~\citep{gunjal2025rubrics,huang2025reinforcement,xu2026alternating}. However, naively generated synthetic rubrics can omit relevant dimensions or contain generic, conflated, or misaligned criteria, reducing judgment accuracy~\citep{shen2026rethinking}. 
One omission from much of the existing literature is evaluating how well a model's synthetic rubrics score responses from future stronger models. In practice, that is how synthetic rubrics would advance the frontier of AI in non-verifiable domains, especially if these rubrics are subsequently edited/validated by experts and must provide lasting value that justifies this effort. 

\paragraph{Rubric construction strategies.} AutoChecklist~\citep{zhou2026autochecklist} considers five paradigms for rubric generation and validation: direct prompting, contrastive generation, inductive extraction from feedback, decomposition of predefined dimensions, and interactive construction using human and LLM feedback. Response-grounded approaches to rubric-generation derive discriminative criteria from observed quality differences between candidate responses~\citep{liu2025openrubrics,zhang2025chasing}. RubricRAG~\citep{dhole2026rubricrag} instead retrieves rubrics for similar queries and uses them as few-shot examples to guide generation.

Our method simply generates an initial draft rubric via a single LLM call, using a prompt that encodes general best practices \citep{rao2026autorubric}. This allows \rubricon{} to be flexibly tailored for various settings where auxiliary inputs can be included in this LLM call, such as when there are expert-written golden responses, unstructured notes on what makes a response good/bad, critiques of particular responses or expert-preferences between them,  
or few-shot examples of expert-crafted rubrics. \rubricon{}'s  refinement stages can be applied to initial draft rubrics generated from any other approach.

\paragraph{Iterative and adaptive refinement.} 
Recursive Rubric Decomposition (RRD)~\citep{shen2026rethinking}
recursively decomposes broad criteria into finer-grained dimensions while filtering misaligned and redundant criteria. Its whitened-uniform (WU) variant further applies correlation-aware weighting of rubric criteria to prevent overlapping criteria from dominating the aggregate score. 
Other refinement approaches include: RubricHub~\citep{li2026rubrichub}, which combines criteria from multiple models and compares high-quality responses; Chasing the Tail~\citep{zhang2025chasing}, which constructs discriminative rubrics from strong, diverse responses to mitigate reward over-optimization; and SibylSense~\citep{xu2026sibylsense},
which adapts a frozen rubric-generator through a memory bank of validated criteria, alternating verifier-guided updates with adversarial generation of candidate responses.

\paragraph{Training rubric generators.} OpenRubrics~\citep{liu2025openrubrics} uses contrastive prompting to derive hard rules and implicit principles from prompts and preference-ordered responses, then fine-tunes a rubric generator and judge on examples filtered for preference consistency. \citet{lv2026learning} train a query-specific rubric generator (QSRG) through reinforcement learning, rewarding human-preference consistency, format validity, and LLM-assessed rubric quality. \citet{sheng2026reinforcing} jointly train one model for reasoning and rubric generation, using evolving criteria for chain-of-thought supervision.

\section{Task Formulation}
\label{subsec:taskform}

We consider open-ended tasks, each specified by a prompt $T$ for which a model should generate a response. 
A rubric for $T$ is a set of weighted binary criteria
$\mathcal{R}=\{(c_k, w_k)\}_{k=1}^{m}$ used to grade any possible response, where each criterion $c_k$ is a
binary (pass/fail) natural-language requirement and $w_k$ encodes
its relative importance. A \emph{verifier} model $v(c_k, y)\in\{0,1\}$ judges whether response
$y$ satisfies each criterion $c_k$. The response $y$ receives an overall normalized score:

\begin{equation}
  s_{\mathcal{R}}(y)
  \;=\;
  \frac{\sum_{k=1}^{m} w_k \, v(c_k, y)}{\sum_{k=1}^{m} \max(w_k, 0)}
  \label{eq:rubricscore}
\end{equation}

We use this score to rank responses from different models or as rewards in post-training. While \rubricon{}'s rubrics employ positive weights for simplicity, some rubrics contain negative-weight criteria (including baseline methods we compare against in our experiments).

A strong rubric satisfies three \emph{desiderata}:
\begin{enumerate}
\item \textbf{Discrimination.} The rubric should assign higher scores to higher-quality responses, clearly separating better from worse.
\item \textbf{Coverage.} The rubric should capture relevant requirements and failure modes that models actually exhibit, so that flawed responses do not receive high scores.
\item \textbf{Headroom.} The rubric should remain sensitive to meaningful quality differences among responses from future models, even as their capabilities improve.
\end{enumerate}

Due to verifier noise and computation costs as well as human-editing/review costs, more criteria are not always desirable in a rubric. A rubric should instead have the fewest criteria that achieve the aforementioned desiderata, and should not include basic criteria that any existing model would pass or subjective/complex criteria that a simple verifier model cannot reliably grade.

\begin{figure*}[t]
  \centering
  \includegraphics[width=\linewidth]{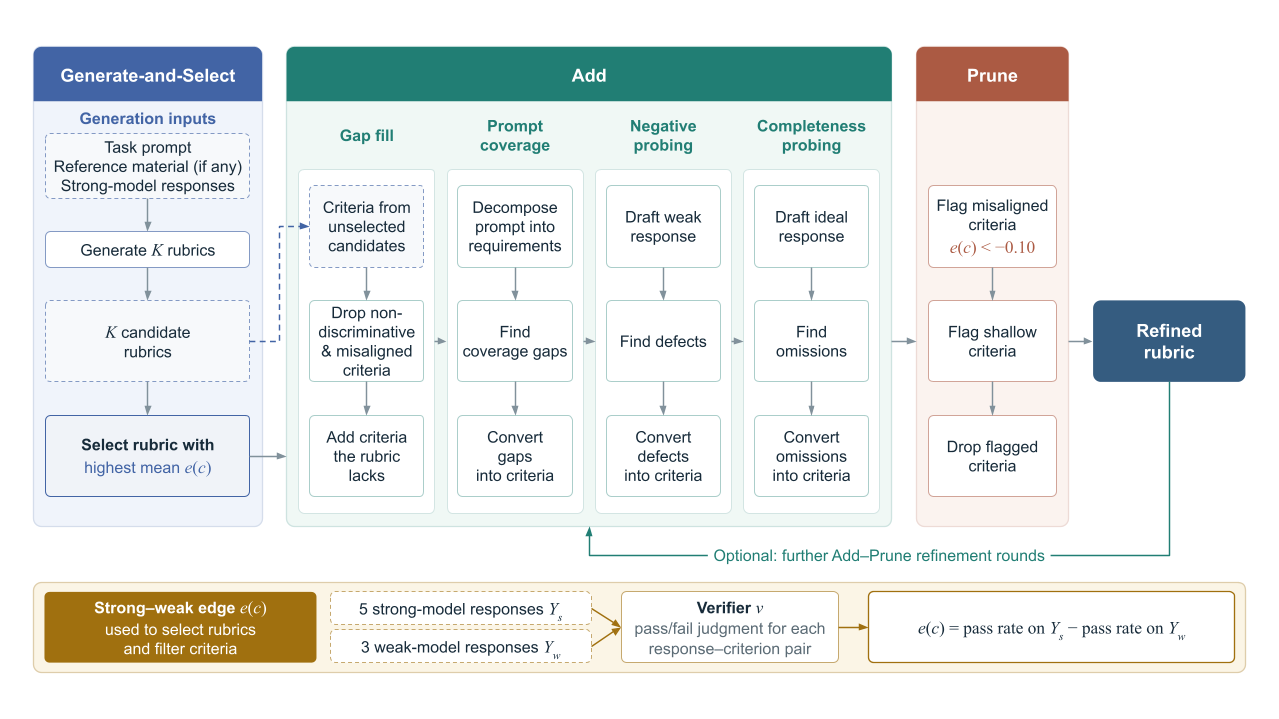}
  \caption{\textbf{The \rubricon{} pipeline.} Given a task prompt, example strong-model responses, and optional reference material, Generate-and-Select drafts $K = 5$ candidate rubrics and selects the rubric with the largest average strong--weak edge across criteria. Add sequentially expands the selected rubric through gap fill, prompt coverage, negative probing, and completeness probing, with each step operating on the rubric from the prior step. The probe steps generate responses intended to satisfy the current rubric, then convert independently identified defects or omissions into new criteria. Prune removes misaligned criteria and those that are satisfiable by evasive or weak responses. The resulting rubric can undergo further Add then Prune refinement rounds.}
  \label{fig:pipeline}
\end{figure*}

\section{Rubricon}

\rubricon{} constructs and refines rubrics through a modular pipeline that can utilize any off-the-shelf LLM as the \emph{generator model}. Unlike rubric generators that require model training, the performance of our approach can be instantly improved whenever better LLMs are released. 
 \rubricon{} can also leverage expert reference responses or expert reasoning as additional context whenever available for a task. It measures criterion pass rates on sample responses and uses LLM-based checks to identify missing requirements and poorly formulated criteria. Its main discriminative signal is the \emph{strong--weak edge}, the gap in criterion pass rates between stronger- and weaker-model responses, adapted from the misalignment filter proposed by \citet{shen2026rethinking}. The pipeline uses these signals to select an initial rubric, add missing criteria, and remove misaligned criteria or criteria that are satisfiable by evasive or weak responses. 

\subsection{Aspects of Effective Rubric Criteria}
\label{subsec:principles}

Beyond the rubric-level desiderata in Section \ref{subsec:taskform}, a good rubric requires well-formed individual criteria. We enumerate a universal set of \textit{Criteria Principles}, a task-agnostic style guide for writing criteria, which we embed into our criterion generation steps. Each criterion should be \textbf{clear and unambiguous}, with an explicit pass condition; \textbf{self-contained}, requiring no external references to assess the response; \textbf{atomic}, testing only one requirement; and \textbf{free of cross-references}, requiring no other criterion or verdict to interpret. These principles govern individual criteria, while the \rubricon{} pipeline in Section \ref{subsec:pipeline} operates on the criterion set. Appendix \ref{app:rubric-school} presents the Criteria Principles guide.

In our rubrics, each criterion contains: \textit{criterion text}, a natural-language requirement that is straightforward to objectively judge as pass or fail; an \textit{importance label}, which determines its \textit{weight} in the aggregate score (Eq. \ref{eq:rubricscore}); and a \textit{category}, such as Factual Accuracy, Reasoning, or Safety. Categories describe the dimension of the response being assessed but do not affect its score according to the rubric.
Our rubric generator assigns each criterion an importance label: Critical, Major, or Minor, for which we assign relative importance  weights of 5, 3, and 1, respectively~\citep{gunjal2025rubrics,bankertoolbench,lommel2024multi,rao2026autorubric}.

\subsection{The \rubricon{} Pipeline}
\label{subsec:pipeline}

\rubricon{} generates and refines rubrics through three stages: \textbf{Generate-and-Select}, \textbf{Add}, and \textbf{Prune} (Figure~\ref{fig:pipeline}). Criteria generated throughout the pipeline follow the Criteria Principles in Section~\ref{subsec:principles}. The three pipeline stages address the rubric-level desiderata in Section \ref{subsec:taskform}: Generate-and-Select targets discrimination, Add expands coverage and helps distinguish among stronger responses, and Prune removes criteria that reward wrong qualities or lead to noisy scores.

Before running \rubricon{} to produce a rubric for a prompt, one first generates a set of example responses to the prompt, some from a known \emph{stronger} model (e.g.\ GPT-6) and some from a known \emph{weaker} model (e.g.\ GPT-4). While one cannot guarantee that the weaker model's response is not better than the stronger model's, appropriately choosing  these LLMs makes this unlikely.

\paragraph{Generate-and-Select.}
We generate $K=5$ candidate rubrics through independent LLM calls for which the input is the task prompt, example responses (from only the strong-model), and any task-provided reference context from experts (Appendix \ref{app:prompt-generation}). Generating multiple rubric candidates broadens coverage and reduces reliance on any single draft, while retaining alternative criteria for the subsequent Add stage. To choose the best initial draft rubric among these candidates, we assess how well their criteria distinguish stronger- from weaker-model responses.

A criterion satisfied by every example response provides no discrimination among those responses, whereas one that weaker-model responses pass more often than stronger-model responses suggests misalignment. Following \citet{shen2026rethinking}, we quantify this using the \emph{strong--weak edge}: the difference in a criterion's pass rates between stronger- and weaker-model responses, for which we first run the verifier to score these example responses against the draft rubric criteria.

Let $Y_s$ and $Y_w$ denote the stronger- and weaker-model response sets, respectively, and let $\bar{v}(c,y)$ denote the verifier pass rate over \(n\) judgments for criterion $c$ against response $y$. We estimate the strong--weak edge as
\begin{equation}
e(c)=\frac{1}{|Y_s|}\sum_{y\in Y_s}\bar{v}(c,y)
-\frac{1}{|Y_w|}\sum_{y\in Y_w}\bar{v}(c,y)
\label{eq:edge}
\end{equation}

A positive edge indicates that stronger-model responses satisfy the criterion more often, suggesting it discriminates in the right direction.

We rely on these edge estimates in both rubric candidate selection and criterion filtering stages. In the selection stage, \rubricon{} retains the candidate rubric with the highest average strong--weak edge, where this average is computed over all criteria in the rubric. The criteria in the unselected rubrics are considered a candidate pool for the subsequent Add stage. 

\paragraph{Add missing coverage.}
Even the selected candidate rubric remains unlikely to satisfy all desiderata in Section \ref{subsec:taskform}. To expand coverage and target headroom, \rubricon{} refines this draft rubric through four complementary addition stages, applied sequentially so that each operates on the rubric augmented by the preceding stages:

\begin{enumerate}
  \item \textbf{Gap fill} retrieves non-redundant criteria from the unselected candidate rubrics to cover expectations missing from the current rubric (Appendix \ref{app:gap-fill}). It first filters the candidate criteria pool such that a criterion is ignored if every example response passes it in every verifier run, or if every strong-model response fails it and every weak-model response passes it in every verifier run. 

  \item \textbf{Prompt coverage} checks whether the rubric captures explicit task requirements. The task's prompt is decomposed into atomic requirements, each marked as covered, partially covered, or uncovered by the current rubric. Uncovered or partially covered requirements are converted into new criteria (Appendix \ref{app:prompt-coverage}).

  \item \textbf{Negative probing} generates adversarially-weak responses (from the same LLM) that are known to be bad but that the current rubric may overlook. A separate LLM reviewer, seeing only the task and one such response, flags concrete defects. A subsequent LLM call converts relevant defects not already covered by the rubric into new criteria (Appendix \ref{app:negative-probing}).

  \item \textbf{Completeness probing} searches for omissions in an intended high-quality response. A model is prompted to generate a response that satisfies every criterion in the current rubric. A separate LLM reviewer, seeing only the task prompt and response, flags missing or incomplete information that a high-quality answer would still need. A final LLM call converts relevant omissions not already covered by the rubric into new criteria (Appendix \ref{app:completeness-probing}).
\end{enumerate}

These four stages expand rubric coverage. Newly generated criteria follow the Criteria Principles and receive importance labels as described in Section~\ref{subsec:principles}. The model is instructed to add only task-relevant criteria that are not already covered in the current rubric and are not overly specific to a particular solution approach. We subsequently use the verifier to score the sample responses against each newly added criterion, allowing misaligned criteria to be identified and removed during pruning.

\paragraph{Prune misleading criteria.}
\label{sec:prune}
Generation and expansion can introduce noisy criteria that degrade response scoring. To support discrimination after expansion, we remove a criterion if its estimated strong--weak edge (Eq.~\ref{eq:edge}) is below a threshold ($-0.10$), or if an LLM assessor flags it as satisfiable by an evasive or weak response (Appendix~\ref{app:prune}). 

The remaining criteria form the refined rubric.
Appendix~\ref{app:evolution} provides examples showing how the Add and Prune stages update a specific rubric's criteria.

\paragraph{Iterative refinement.} 
The Add and Prune stages can be repeatedly iterated. We refer to one application of Add followed by Prune as a refinement round, with each subsequent round starting from the preceding round’s refined rubric. Gap fill continues to draw from the original unselected drafts, while coverage checks and probes are rerun against the updated rubric. The same strong- and weak-model responses are reused to score and prune criteria across rounds. 
After a second refinement round, we consistently observe higher alignment with expert-authored rubrics (Section~\ref{subsec:iteration}).

\section{Experimental Setup}
\label{sec:experimental-design}

\subsection{Assessing Rubric Quality}
\label{sec:setup-metrics}
We assess the quality of each generated rubric by how well the resulting rubric scores 
align with expert human judgments of the same responses.

\paragraph{Alignment with expert-authored rubrics.}
Human-written criteria describe what experts expect from a good response and can rank responses by how well they meet those expectations. Agreement with these rankings tests whether generated rubrics capture similar quality distinctions, even if the rubrics assign different numerical scores. 
To quantify this alignment, we use the verifier to score a set of \emph{assessment responses} from many models against both auto-generated and expert-authored rubrics, then rank the responses according to each rubric, and finally report Spearman's rank correlation ($\rho$) between these  two rankings. The overall alignment for a dataset is the average $\rho$ across all of its tasks.

\paragraph{Alignment with human preferences.}
Human pairwise preferences between two responses provide an alternative measure of quality, albeit often coarser/noisier than rubric-based scoring \citep{rezaei2026online,li-etal-2024-dissecting}. Alignment with these preferences indicates how well rubric scores reflect human assessments of relative response quality. 
We report \textit{selective preference accuracy}: after scoring all responses against the generated rubric, we retain response-pairs whose normalized rubric scores differ by more than a fixed margin ($\delta=0.10$) and measure how often the higher-scoring response matches the human preference. We use this margin to avoid treating small score differences that may arise from verifier variability as meaningful preferences. 

\subsection{Verifier Scoring}
\label{sec:verifier}

Our verifier $v(c,y)\in\{0,1\}$ is an LLM that judges whether a response $y$ satisfies a criterion $c$. As in ~\citet{profbench}, the verifier receives only the response and a single criterion, without the original task prompt, under the assumption that criteria are self-contained and can be judged independently. To account for variability in its judgments, we evaluate each response-criterion pair $n$ times using the prompt in Appendix \ref{app:verifier-prompt}, denoting the verdict from the $j$-th evaluation by $v_j(c,y)$. Appendix~\ref{app:verifier} shows that our verifier exhibits high agreement with expert judgments when grading the same responses against the same rubrics.

\paragraph{Verifier scoring within the \rubricon{} pipeline.}

For rubric selection and pruning, we sample five responses from a stronger model and three from a weaker model per task. Each response-criterion pair receives $n=3$ verifier evaluations. We average all verifier judgments of each model's example responses to obtain its pass rate for a criterion, then subtract the weaker model's pass rate from the stronger model's to estimate the strong--weak edge defined in Eq.~\ref{eq:edge}.

\paragraph{Verifier scoring for rubric-quality assessment.}

For assessing rubric quality, we apply the verifier to completed rubrics and fixed responses separate from those used during rubric construction. The verifier produces $n=3$ judgments for each response-criterion pair. We aggregate the $j$-th judgment across all criteria to obtain a response score for scoring run $j$, using Eq. \ref{eq:rubricscore}.

For \textit{alignment with expert-authored rubrics}, we use both automatically-generated and human-written rubrics to score the same set of \textit{assessment responses} from various LLMs. Within each scoring run, we compute $\rho$ between the two sets of response scores, yielding $n$ correlation estimates per task. We report the mean correlation across all tasks and scoring runs within each benchmark. For \textit{selective preference accuracy}, we apply the margin to each of the $n$ score comparisons for a task, then pool the retained comparisons across all tasks to compute accuracy.

\subsection{Datasets}
\label{sec:setup-benchmarks}
We evaluate the generated rubrics on five benchmarks spanning professional, consumer, clinical, and general-purpose tasks (Table \ref{tab:benchmarks}). We use expert-authored rubrics from ProfBench~\citep{profbench}, PRBench~\citep{prbench}, ACE~\citep{ace}, and HealthBench~\citep{arora2025healthbench} to measure alignment between rankings of different LLM responses produced by generated v.s.\ expert-authored rubrics. We use human preference labels from
RubricBench~\citep{rubricbench} to measure human preference agreement. Dataset descriptions, preprocessing, reference material, and final task counts are detailed in Appendix~\ref{app:datasets}.

\subsection{Baselines}
\label{sec:baselines}

We compare \rubricon{} against two families of baselines described in Section \ref{rel_works}: prompt-based methods, including an Agentic baseline (Appendix~\ref{app:agentic-prompt}) and RRD$_{\mathrm{WU}}$~\citep{shen2026rethinking}, and post-trained rubric generators, including the Query-Specific Rubric Generator (QSRG)~\citep{lv2026learning} and OpenRubrics~\citep{liu2025openrubrics}. \rubricon{} belongs to the prompt-based family. Each method produces one rubric per task, and across all five benchmarks, responses are graded against these rubrics using the same verifier and $n=3$ replicate scoring runs (Section \ref{sec:verifier}).

The \emph{Agentic} baseline runs in a Codex SDK \citep{openai2026codexsdk} harness with full local tool access and web access disabled, freely choosing its rubric-construction procedure. It tests whether an unconstrained agent can match \rubricon{}'s structured pipeline. Appendix~\ref{app:agentic-prompt} provides details. 
Both the Agentic and RRD baselines receive the same task prompt, example model responses, and available reference material as \rubricon{} (Section
\ref{sec:setup-benchmarks}).

The post-trained rubric generators, OpenRubrics and QSRG, are run as published and receive the task prompt alone.
We also examine the effect of input context through QSRG+Context, which uses the same trained generator as query-only QSRG, but augments the task query with example model responses and any dataset-provided reference material (Appendix~\ref{app:datasets}). 
The post-trained rubric generators use their respective models: QSRG and QSRG+Context use a Qwen3-30B-A3B-based generator, while OpenRubrics uses Rubric-RM-8B, based on Qwen3-8B \citep{yang2025qwen3}.

\subsection{Models Employed}
\label{sec:models-employed}

For experiments measuring alignment with human rubrics and human preferences, all prompt-based rubric generation methods, including \rubricon{}, are powered by the same generator model: GPT-5.1. 

For rubric generation methods that leverage example LLM responses during rubric construction, we use responses from GPT-5.1 and Llama-3.1-8B-Instruct \citep{grattafiori2024llama} as the \emph{strong} and \emph{weak} models, respectively. 
We use GPT-5 mini as the verifier model for criterion grading throughout, because this model was observed to provide a favorable accuracy-cost tradeoff, with strong agreement with human expert judgments on the same responses and rubrics (Appendix~\ref{app:verifier-human}).

For our experiments measuring alignment with human rubrics, we use generated rubrics to score a separate set of responses from the following models: GPT-5.2 Instant, GPT-5.2 Thinking, Claude Sonnet 4.6, Claude Opus 4.6, Gemini 2.0 Flash-Lite, Gemini 2.5 Flash, Gemini 3 Flash, and Gemini 3.1 Pro. We generate one response from each model per task, yielding eight responses to score with each rubric (not used during rubric construction).    Appendix~\ref{app:inference-settings} reports inference settings.

% \paragraph{Measuring headroom for newer models.} 
Some of these responses come from newer/stronger models than those used for rubric generation and verification. 
This reflects the fact that good rubrics must remain useful for evaluating responses from future models, which remains underexplored in prior work. \citet{shen2026rethinking}, \citet{xu2026sibylsense}, and \citet{zhang2025chasing} evaluate their generated rubrics mainly on responses from models weaker than or comparable to the rubric generator, although some of them leverage responses from stronger models during rubric construction.

\section{Results for LLM Evaluation}
\label{sec:full_results}

Unless otherwise specified, \rubricon{} results are for the rubric obtained after two refinement rounds.

\subsection{Alignment with Human Rubrics}

\rubricon{} achieves the highest $\rho$ on all four benchmarks in Table~\ref{tab:summary_main}. It attains a macro-average $\rho$ of $0.428$ across these datasets, representing a $9.0\%$ relative improvement over RRD$_{\mathrm{WU}}$, the strongest alternative approach. On ACE, \rubricon{} exceeds the next-best method by $0.052$, its largest margin across the four benchmarks. Additional context improves QSRG on three benchmarks, but overall both post-trained rubric generators produce substantially less expert-aligned rubrics than  \rubricon{}.

\begin{table*}[t]
\centering
\scriptsize
\setlength{\tabcolsep}{3pt}
\renewcommand{\arraystretch}{1.12}
\resizebox{\textwidth}{!}{%
\begin{tabular}{lccccc}
\toprule
\textbf{Method} & \textbf{ProfBench} & \textbf{ACE} & \textbf{PRBench} & \textbf{HealthBench} & \textbf{RubricBench} \\
{\color{black!60}Metric} & {\color{black!60} $\rho$} & {\color{black!60} $\rho$} & {\color{black!60} $\rho$} & {\color{black!60} $\rho$} & {\color{black!60}Accuracy (\%)} \\
\midrule
QSRG & $0.361 \pm 0.082$ & $0.225 \pm 0.043$ & $0.226 \pm 0.013$ & $0.172 \pm 0.008$ & $52.9 \pm 1.6$ \\
QSRG+Context & $0.267 \pm 0.070$ & $0.301 \pm 0.038$ & $0.353 \pm 0.012$ & $0.279 \pm 0.009$ & $52.4 \pm 1.6$ \\
OpenRubrics & $0.202 \pm 0.068$ & $0.276 \pm 0.038$ & $0.071 \pm 0.014$ & $0.119 \pm 0.010$ & $57.5 \pm 1.7$ \\
Agentic & $0.283 \pm 0.063$ & $0.333 \pm 0.039$ & $0.363 \pm 0.014$ & $0.294 \pm 0.009$ & $59.3 \pm 1.9$ \\
RRD$_{\mathrm{WU}}$ & $0.394 \pm 0.076$ & $0.316 \pm 0.040$ & $0.497 \pm 0.011$ & $0.365 \pm 0.009$ & $58.7 \pm 1.7$ \\
\midrule
\rubricon{} & $\mathbf{0.424 \pm 0.083}$ & $\mathbf{0.385 \pm 0.035}$ & $\mathbf{0.516 \pm 0.011}$ & $\mathbf{0.388 \pm 0.009}$ & $\mathbf{59.8 \pm 1.9}$ \\
\bottomrule
\end{tabular}%
}
\caption{\textbf{Alignment with human rubrics and human preferences.} \rubricon{} achieves the highest observed scores across all five benchmarks. Across datasets, it improves macro-average $\rho$ by 9.0\% relative to the strongest baseline, RRD$_{\mathrm{WU}}$ ($0.428$ vs.\ $0.393$), and selective preference accuracy by 0.5 percentage points over the strongest preference baseline, Agentic ($59.8\%$ vs.\ $59.3\%$). The first four benchmarks report $\rho$ between scores under generated and expert-authored rubrics. RubricBench reports selective preference accuracy (\%), evaluated on response pairs whose normalized rubric scores differ by more than $0.10$ (retained counts in Table \ref{tab:rubricbench_results}). The $\pm$ values indicate standard errors estimated by bootstrapping tasks ($B=500$ resamples).}
\label{tab:summary_main}
\end{table*}

\subsection{Alignment with Human Preference}

\rubricon{} achieves the highest observed selective preference accuracy, exceeding the next-best method, the Agentic baseline, by $0.5$ percentage points (rightmost column in Table~\ref{tab:summary_main}). 
Note that the alignment with human preferences is fairly low for all methods, because \citet{rubricbench} purposefully populated RubricBench with challenging examples.

\subsection{Iterative Refinement Across Multiple Rounds}
\label{subsec:iteration}
\rubricon{} is an iterative pipeline: the Add and Prune stages can be reapplied to refine the rubric produced by the preceding round.  Table~\ref{tab:rc_second_round} shows that the first refinement round yields an $11.9\%$ relative improvement in macro-average $\rho$ over the selected initial rubric. The second round yields a further $5.1\%$ relative improvement over the first round. Compared with RRD$_{\mathrm{WU}}$, the relative improvements after one and two rounds are $3.7\%$ and $9.0\%$, respectively.

\subsection{Performance Across Refinement Steps}
\label{subsec:ablation}

We examine the contribution of each pipeline stage by assessing the quality of intermediate rubrics produced during the first refinement round. Table~\ref{tab:rc_stage_macro} summarizes results across the four benchmarks with expert-authored rubrics. Detailed per-benchmark results appear in Table~\ref{tab:rc_stage_results}. The results show that each stage of the \rubricon{} pipeline is valuable: every stage improves alignment on at least one benchmark, and most stages also improve the macro-average $\rho$ across the four benchmarks. These results suggest that different stages are useful for different types of tasks, which could differ substantially in their expert-authored rubrics and what constitutes a good response.

Some steps in the Add stage slightly reduce macro-average $\rho$ while improving this alignment with expert-rubrics on individual benchmarks. Expanding rubric coverage can introduce noisy or misleading criteria alongside useful ones. Subsequent pruning is designed to remove these problematic criteria and proves empirically effective overall. The improvement in alignment after pruning suggests that coverage expansion and filtering play complementary roles, even when gains are not observed immediately after the Add step.

\begin{table}[htbp]
\centering
\footnotesize
\setlength{\tabcolsep}{5pt}
\renewcommand{\arraystretch}{1.05}
\begin{tabular}{lcc}
\toprule
\textbf{Stage} & \shortstack{\textbf{Macro-avg.}\\\textbf{$\rho$}} & \shortstack{\textbf{Avg. Rubric}\\\textbf{Size}} \\
\midrule
Generation & 0.359 & 20.6 \\
Selected Rubric & 0.364 & 20.6 \\
Gap Fill & 0.362 & 23.9 \\
Prompt Coverage & 0.362 & 24.6 \\
Negative Probing & 0.388 & 28.3 \\
Completeness Probing  & 0.395 & 31.0 \\
Prune & 0.408 & 26.7 \\
\bottomrule
\end{tabular}
\caption{Alignment with expert-authored rubrics (macro-averaged $\rho$ across datasets) and average rubric size after each stage of the first refinement round. Rows show cumulative refinement. Values in the top row are averaged over $K= 5$ independently generated rubrics.}
\label{tab:rc_stage_macro}
\end{table}

\section{Using Rubrics for Post-Training}
\label{sec:posttraining}

Beyond model evaluation, rubrics are also used to compute rewards for post-training LLMs via reinforcement learning. Here we assess how effective \rubricon{} rubrics are for RL post-training of the Qwen3.5-9B model \citep{qwen2026qwen35}. Our experiment trains the same model using the same recipe against rubrics from different generators, and then measures the quality of each resulting model based on how it scores against expert-authored rubrics for held-out tasks.

\paragraph{Data.}
We post-train models on the HealthBench dataset, randomly splitting its 1,704 prompts into 1,204 training prompts and 500 held-out evaluation prompts.

\paragraph{Training conditions.}
We compare two training-time reward sources: RRD$_{\mathrm{WU}}$, our strongest baseline for alignment with expert-authored rubrics, and \rubricon{} rubrics obtained after two refinement rounds. We fine-tune the released Qwen3.5-9B model checkpoint with GRPO \citep{shao2024deepseekmath} for eight epochs. Training configurations are detailed in Appendix~\ref{app:training-details}, Table~\ref{tab:training-config}. Apart from the training-time rubrics, the model, training recipe, data split, and evaluation procedure are held fixed.

\paragraph{Held-out policy evaluation.}
At each evaluated model checkpoint, we sample three independent responses per held-out prompt and score them against physician-authored HealthBench criteria (using the same verifier). We report \textit{avg@3}, the average of the (expert-authored) rubric scores of these three independently sampled responses, subsequently averaged across all held-out evaluation prompts. We also report paired 95\% bootstrap confidence intervals for the difference in mean scores between models trained using different rubrics.

\begin{figure}[!t]
  \centering
  \includegraphics[width=\linewidth]{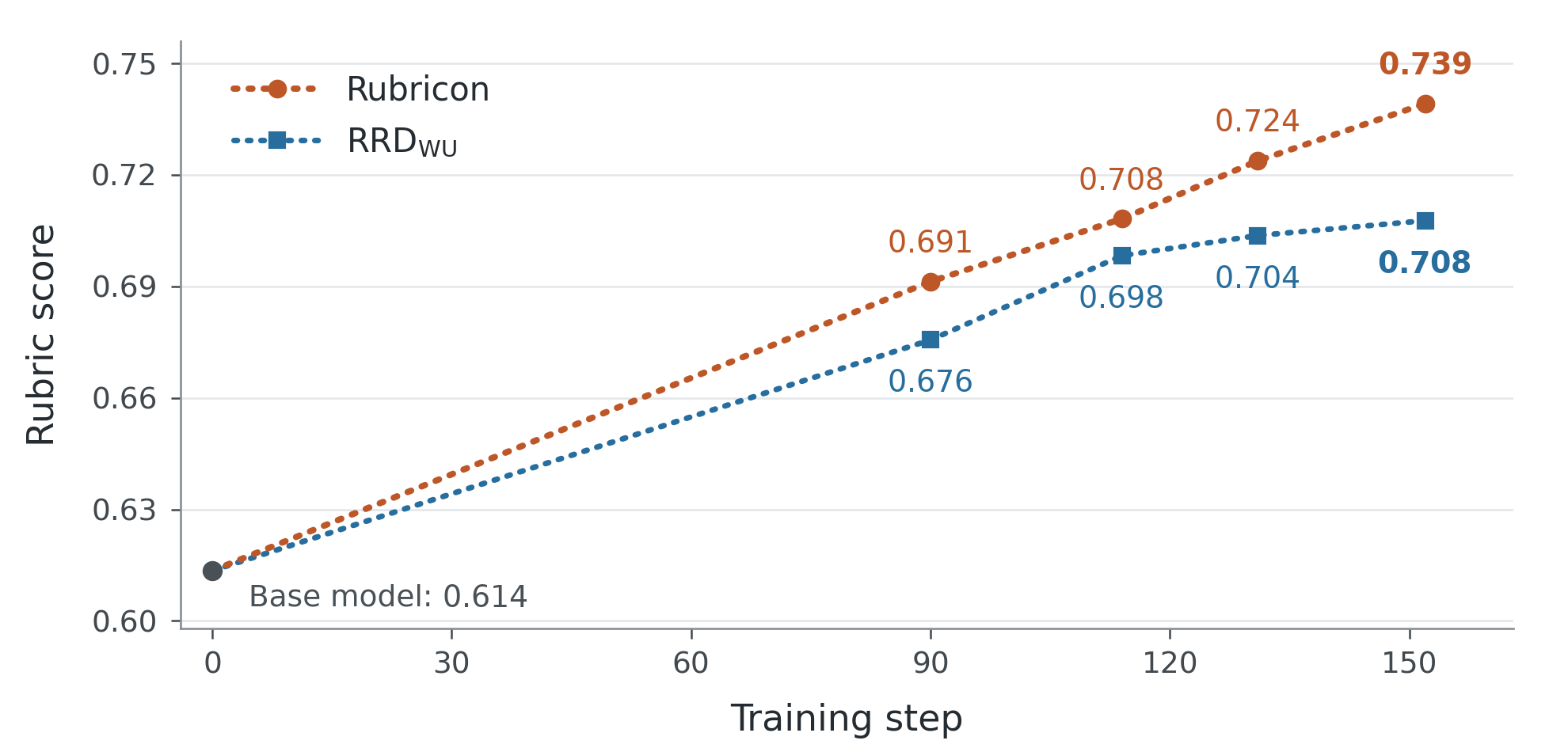}
  \caption{\textbf{\rubricon{} rewards translate into stronger held-out post-training performance.} Qwen3.5-9B trained with \rubricon{} rubric rewards outperforms the same model trained with RRD$_{\mathrm{WU}}$ rewards at every evaluated post-training checkpoint, reaching a 4.5\% higher score at the last evaluated checkpoint. Scores are averaged over three independently sampled responses per prompt, then across 500 held-out HealthBench prompts. The \rubricon{}-trained model achieves a 20.5\% higher held-out score than the starting model checkpoint.}
\label{fig:healthbench-rubric-source}
\end{figure}

\paragraph{Results.} 
Under the fixed recipe, training with \rubricon{} rubrics yields higher held-out HealthBench scores than training with RRD$_{\mathrm{WU}}$ rubrics at every evaluated checkpoint (Figure~\ref{fig:healthbench-rubric-source}). The final-checkpoint improvement is $0.0315$ (paired 95\% bootstrap CI: $[+0.0207,+0.0427]$). 
Here we use Gemini 3.5 Flash-Lite as the verifier model for both rollout scoring during training and held-out evaluation graded against expert-authored rubrics. 
The advantage persists when the same final-checkpoint responses are re-scored by GPT-5 mini or Claude Sonnet 5 as verifiers instead (Appendix~\ref{app:verifier-sensitivity-posttraining}). These results support \rubricon{}'s efficacy for providing reward signals in post-training.

\section{Discussion}

Our results show that rubric quality can be automatically improved beyond basic LLM or agentic generation, without requiring additional data collection/annotation or model fine-tuning. 
By systematically selecting discriminative drafts, filling gaps in coverage, and removing misleading criteria, \rubricon{} produces more effective rubrics for LLM evaluation and training. The key is a methodical test-time scaling strategy that does not purely rely on LLM self-critique and ensembling, but additionally grounds refinement stages in how draft rubrics actually score example responses.

\section*{Limitations}

Our experiments are limited to single-turn LLM responses in certain domains and to the English language. It remains unclear whether the rubrics generated in these experiments will continue to distinguish response quality as models continue to improve beyond the most advanced LLMs considered in our experiments. As models are increasingly trained as agentic  and multi-agent systems, an agentic or multi-agent approach to rubric generation may one day surpass \rubricon{}.

\rubricon{} relies heavily on model-derived signals and the resulting automatically generated rubrics are far from perfect. Because it is compatible with any LLM, \rubricon{} rubrics will improve as models improve, but the flaws in model responses will also get subtler. In these non-verifiable domains where quality judgments are driven by expert taste, one cannot rely on theoretical guarantees because what constitutes a good rubric can greatly vary from one dataset to another.

Thus we recommend always having human experts review and improve the rubrics automatically generated by methods like \rubricon{}, which still yields enormous time-savings compared to manually crafting rubrics from scratch. 
A beneficial property of our method is that its modular stages are easy to adjust based on expert feedback, and a prompt-based approach like ours can easily incorporate expert suggestions/conventions as well (in contrast to a trained rubric generator). \rubricon{}'s refinement stages can be seamlessly iterated with manual expert updates to rubrics, or even applied to human-crafted initial rubrics. Optimal human-AI collaboration to construct LLM evaluations effectively and affordably remains an important topic for future research.

\clearpage 
\bibliographystyle{abbrvnat}
\bibliography{bibliography}

\clearpage 
\appendix

\begin{center}
\huge \textbf{Appendix}
\end{center}

\section{Benchmark Dataset Details}
\label{app:datasets}

\subsection{Benchmarks for Assessing Rubric Quality}
\paragraph{Benchmarks with human rubrics.} 

To ensure they are high-quality, we first reviewed many expert-authored rubrics in every dataset used in our experiments on alignment against expert rubrics. Many other datasets were omitted from consideration during these reviews.

We use the Finance, Consulting, and Physics domains of ProfBench~\citep{profbench} and the  Finance and Law tasks of PRBench~\citep{prbench}. Both benchmarks contain tasks requiring professional knowledge and reasoning. We also use the open development split of the AI Consumer Index (ACE)~\citep{ace}, which covers everyday consumer tasks in Shopping, DIY, Gaming, and Food, and HealthBench~\citep{arora2025healthbench}, which contains health conversations with physician-authored, importance-weighted rubrics. On these four benchmarks, we score the same responses using generated and expert-authored rubrics and report rank correlations as defined in Section~\ref{sec:setup-metrics}. HealthBench is also used for post-training experiments because of its size and prior work using generated rubrics as reinforcement-learning
rewards~\citep{shen2026rethinking}.

\paragraph{Benchmark with human preferences.} We use the released RubricBench preference set~\citep{rubricbench}, comprising 1,147 pairwise preference examples drawn from datasets like HelpSteer3, PPE, and RewardBench2. Each example contains a task, two LLM responses, and a binary human preference label. We score each response independently using the generated rubric and report selective preference accuracy defined in Section \ref{sec:setup-metrics}.

\subsection{Reference Material for Rubric Construction}
\rubricon{}, Agentic, RRD$_{\mathrm{WU}}$, and QSRG+Context are provided with task prompts, example model responses, and any auxiliary per-task reference material in each dataset (see Appendix \ref{app:prompt-generation}).  This reference material includes expert-authored step-by-step reasoning for PRBench and expert-authored example responses for HealthBench. 
We treat the human-written responses in HealthBench as  example responses rather than golden reference completions, after finding that they do not score highly under its own expert-authored rubrics. 
No additional reference material is used for ProfBench, ACE, or RubricBench. QSRG and OpenRubrics receive the task prompt alone. The expert-authored rubrics used for alignment-evaluation are never provided to the rubric-generators.

\subsection{Data Preprocessing} 
We filter our datasets to the subset of examples for which we are more confident in the quality of expert-authored rubrics.
We require at least 10 expert-authored criteria for ProfBench, PRBench, and HealthBench to support graded, discriminative scoring and reduce ties. We do not apply the 10-criterion minimum to ACE, whose human-authored rubrics contain 7 criteria on average. We restrict PRBench and HealthBench to single-turn tasks and require expert-written responses as reference material for HealthBench. We exclude ProfBench's Chemistry domain after identifying issues in the expert-authored rubrics, such as vague or hedged criteria and inconsistent numerical-precision requirements. Table~\ref{tab:benchmarks} reports final task counts and the reference material used for each dataset.

\begin{table*}[t]
\centering
\small
\begin{tabular}{lllrl}
\toprule
\textbf{Benchmark} & \textbf{Domain(s)} & \textbf{Human annotations} & \textbf{Tasks} & \textbf{Ref.\ material} \\
\midrule
ProfBench   & Finance, Consulting, Physics    & Human rubrics & 28      & --- \\
PRBench     & Finance, Law                    & Human rubrics & 656     & expert step-by-step reasoning \\
ACE         & Shopping, DIY, Gaming, Food     & Human rubrics & 80      & --- \\
HealthBench & Health                          & Human rubrics & 1{,}704 & example response \\
RubricBench & Chat, IF, STEM, Code, Safety & Human preferences  & 1{,}147     & --- \\
\bottomrule
\end{tabular}
\caption{Evaluation benchmarks. \textit{Human annotations} specifies how generated rubrics are assessed: alignment with scores from human-authored rubrics or human preferences between responses. \textit{Tasks} gives the number of tasks evaluated after preprocessing. \textit{Ref. material} lists the inputs provided for each task in the rubric generation prompt (see Appendix \ref{app:prompt-generation}).}
\label{tab:benchmarks}
\end{table*}

\section{Details about the Verifier Model}
\label{app:verifier}

\begingroup
\let\sectionsBCTable\table
\expandafter\let\expandafter\sectionsBCTableStar\csname table*\endcsname
\renewcommand{\table}[1][]{\sectionsBCTable[!htbp]}
\expandafter\renewcommand\csname table*\endcsname[1][]{\sectionsBCTableStar[!htbp]}

\subsection{Agreement with Expert Human Graders}
\label{app:verifier-human}
\begin{table*}[t]
\centering
\small
\begin{tabular}{lcccc}
\toprule
& \multicolumn{2}{c}{Cohen's $\kappa$} & \multicolumn{2}{c}{F1} \\
\cmidrule(lr){2-3}\cmidrule(lr){4-5}
Verifier Model & Our Prompt & ProfBench & Our Prompt & ProfBench \\
\midrule
GPT-5 mini & \textbf{0.581} & 0.549 & \textbf{0.802} & 0.786 \\
GPT-5.2 & 0.560 & 0.556 & 0.777 & 0.777 \\
Claude Sonnet 4.6 & 0.556 & 0.533 & 0.785 & 0.760 \\
Gemini 3.1 Pro & 0.580 & \textbf{0.582} & 0.791 & \textbf{0.794} \\
\bottomrule
\end{tabular}
\par\medskip
\caption{Using our verifier prompt, GPT-5 mini achieves the highest observed Cohen's $\kappa$ and F1 among the tested verifiers. We report criterion-level agreement (Cohen's $\kappa$ and F1) with expert annotations on ProfBench using our judge prompt and the original ProfBench judge prompt. On a 367-criterion subset that we independently re-annotated, (non-expert) human agreement with the ProfBench expert annotations is $\kappa = 0.281$ (64.0\% raw agreement). The GPT-5 mini verifier reaches $\kappa = 0.455$ (72.5\% raw agreement) against the expert labels on the same subset.}
\label{tab:verifier}
\end{table*}

Rubric-based evaluation depends on the verifier's ability to correctly judge whether responses satisfy each criterion. To assess the quality of the verifier used in our experiments, we compare LLM verifier judgments with expert human annotations on over 3,400 response-criterion pairs from ProfBench \citep{profbench}. For each pair, the verifier is assessing the same response against the same expert-authored criterion, with the purpose of measuring the criterion-level grading accuracy, separate from the quality of generated rubrics.

We evaluate four verifier models with two prompts: our verifier prompt (Appendix \ref{app:verifier-prompt}), which we use for all our experiments, and the verifier prompt used by ProfBench \citep{profbench}. Our prompt generally yields higher agreement and F1 scores than the ProfBench prompt (Table \ref{tab:verifier}). GPT-5 mini achieves the best performance while remaining cost-effective and highly stable. Across 8 repeated runs, GPT-5 mini's judgments agree on 97.2\% of criteria (mean pairwise $\kappa = 0.943$).

\subsection{Post-training Results Across Verifiers}
\label{app:verifier-sensitivity-posttraining}

To further validate the post-training results with alternative verifiers, we re-scored the final-checkpoint responses with GPT-5 mini and Claude Sonnet 5 on the same physician-authored HealthBench rubrics. The results in Table~\ref{tab:healthbench-verifier-sensitivity} confirm that \rubricon{}'s advantage over RRD$_{\mathrm{WU}}$ holds across all three verifier models.

\begin{table*}[t]
\centering
\begin{tabular}{lccc}
\toprule
Verifier Model & RRD$_{\mathrm{WU}}$ & \rubricon{}
& $\Delta \pm \mathrm{SE}_{\mathrm{paired}}$ \\
\midrule
Gemini 3.5 Flash-Lite & 0.7077 & \textbf{0.7392} & $0.0315 \pm 0.0057$ \\
GPT-5 mini           & 0.6482 & \textbf{0.6753} & $0.0271 \pm 0.0052$ \\
Claude Sonnet 5      & 0.5357 & \textbf{0.5710} & $0.0353 \pm 0.0059$ \\
\bottomrule
\end{tabular}
\caption{Verifier sensitivity on held-out HealthBench evaluation dataset. Each verifier scores the same saved responses from the two policies
after eight training epochs against physician-authored rubrics.
$\Delta$ denotes \rubricon{} minus RRD$_{\mathrm{WU}}$. Standard errors are computed across paired prompt-level differences.}
\label{tab:healthbench-verifier-sensitivity}
\end{table*}

\subsection{Verifier Prompt}
\label{app:verifier-prompt}
\promptfile{Verifier prompt}{artifacts/prompts/VERIFIER_PROMPT.txt}

\section{Additional Results and Analysis}

\subsection{Rubric Sizes and Retained Human Preference Comparisons}
Tables~\ref{tab:rubric_size_comparison} and~\ref{tab:rubricbench_results} complement the main results (Table~\ref{tab:summary_main}) with average rubric sizes across methods and benchmarks. Table~\ref{tab:rubricbench_results} also reports the number of retained comparisons for the RubricBench human-preference evaluation. 

\begin{table*}
\centering
\small
\setlength{\tabcolsep}{6pt}
\renewcommand{\arraystretch}{1.15}
\begin{tabular}{lrrrr}
\toprule
\textbf{Method} & \textbf{ProfBench} & \textbf{ACE} & \textbf{PRBench} & \textbf{HealthBench} \\
\midrule
Expert-authored & 30.5 & 7.4 & 16.6 & 14.7 \\
\midrule
QSRG & 22.1 & 19.5 & 20.3 & 19.9 \\
QSRG+Context & 21.2 & 20.3 & 20.6 & 20.1 \\
OpenRubrics & 11.8 & 9.3 & 9.3 & 7.9 \\
Agentic & 33.9 & 24.3 & 28.4 & 24.1 \\
RRD$_{\mathrm{WU}}$ & 26.0 & 26.7 & 25.4 & 23.4 \\
\midrule
\rubricon{} & 41.9 & 25.0 & 38.8 & 23.4 \\
\bottomrule
\end{tabular}
\caption{Average number of criteria per rubric on the four datasets where we report alignment against human-expert rubrics. The average size of these expert-authored rubrics is included for reference.}
\label{tab:rubric_size_comparison}
\end{table*}

\begin{table}
\centering
\small
\setlength{\tabcolsep}{4pt}
\renewcommand{\arraystretch}{1.15}
\begin{tabular}{lcc}
\toprule
\textbf{Method}
& {\color{black!60}\shortstack{Retained\\comparisons}}
& {\color{black!60}\shortstack{Avg. rubric\\size}} \\
\midrule
QSRG & 1{,}966 & 20.0 \\
QSRG+Context & 2{,}046 & 20.1 \\
OpenRubrics & 2{,}133 & 7.2 \\
Agentic & 1{,}576 & 20.0 \\
RRD$_{\mathrm{WU}}$ & 1{,}938 & 20.5 \\
\midrule
\rubricon{} & 1{,}760 & 17.3 \\
\bottomrule
\end{tabular}
\caption{Number of retained comparisons and average rubric size on RubricBench~\citep{rubricbench}. We retain comparisons with an absolute rubric score difference above $0.10$ (Section~\ref{sec:setup-metrics}). Counts include three scoring runs per task, giving 3,441 comparisons across 1,147 tasks before filtering.}
\label{tab:rubricbench_results}
\end{table}

\subsection{Detailed Performance Across Refinement Steps}
\label{app:ablation}
Table~\ref{tab:rc_stage_results} provides per-benchmark $\rho$ and rubric sizes after each step of the first refinement round, supplementing Section~\ref{subsec:ablation}. Gains vary across benchmarks, with no single stage consistently providing the largest improvement.

\begin{table*}[htbp]
\centering
\scriptsize
\setlength{\tabcolsep}{6pt}
\renewcommand{\arraystretch}{1.2}
\resizebox{\textwidth}{!}{%
\begin{tabular}{lcccc}
\toprule
\textbf{Stage}
& \textbf{ProfBench}
& \textbf{ACE}
& \textbf{PRBench}
& \textbf{HealthBench} \\
\midrule
Selected Rubric 
& $0.345 \pm 0.074$
& $0.352 \pm 0.037$
& $0.421 \pm 0.012$
& $0.339 \pm 0.009$ \\
After Add Stage (round 1)
& $0.358 \pm 0.081$
& $0.376 \pm 0.037$
& $0.486 \pm 0.012$
& $0.360 \pm 0.009$ \\
After Prune Stage (round 1)
& $0.405 \pm 0.084$
& $0.354 \pm 0.037$
& $0.500 \pm 0.012$
& $0.371 \pm 0.009$ \\
After Add Stage (round 2)
& $0.374 \pm 0.086$
& $0.354 \pm 0.036$
& $0.500 \pm 0.012$
& $0.372 \pm 0.009$ \\
After Prune Stage (round 2)
& $\mathbf{0.424 \pm 0.083}$
& $\mathbf{0.385 \pm 0.035}$
& $\mathbf{0.516 \pm 0.011}$
& $\mathbf{0.388 \pm 0.009}$ \\
\bottomrule
\end{tabular}%
}
\caption{Iterative refinement improves alignment on every benchmark. A second refinement round increases macro-average $\rho$ by 17.6\% over the selected draft rubric. Rows are cumulative, with each stage operating on the preceding rubric. Results are reported as $\rho$ ± task-bootstrap standard error.}
\label{tab:rc_second_round}
\end{table*}

\begin{table*}[htbp]
\centering
\scriptsize
\setlength{\tabcolsep}{3pt}
\renewcommand{\arraystretch}{1.2}
\resizebox{\textwidth}{!}{%
\begin{tabular}{lcccccccc}
\toprule
\textbf{Method / stage} & \multicolumn{2}{c}{\textbf{ProfBench}} & \multicolumn{2}{c}{\textbf{ACE}} & \multicolumn{2}{c}{\textbf{PRBench}} & \multicolumn{2}{c}{\textbf{HealthBench}} \\
\cmidrule(lr){2-3}\cmidrule(lr){4-5}\cmidrule(lr){6-7}\cmidrule(lr){8-9}
 & {\color{black!60}$\rho$} & {\color{black!60}Size} & {\color{black!60}$\rho$} & {\color{black!60}Size} & {\color{black!60}$\rho$} & {\color{black!60}Size} & {\color{black!60}$\rho$} & {\color{black!60}Size} \\
\midrule
Selected Rubric & $0.345 \pm 0.074$ & 23.6 & $0.352 \pm 0.037$ & 19.4 & $0.421 \pm 0.012$ & 21.3 & $0.339 \pm 0.009$ & 17.9 \\
Gap Fill & $0.327 \pm 0.077$ & 27.0 & $0.364 \pm 0.039$ & 22.8 & $0.417 \pm 0.012$ & 24.8 & $0.340 \pm 0.009$ & 21.0 \\
Prompt Coverage & $0.323 \pm 0.080$ & 29.0 & $0.368 \pm 0.038$ & 23.1 & $0.417 \pm 0.013$ & 25.0 & $0.339 \pm 0.009$ & 21.1 \\
Negative Probing & $0.346 \pm 0.077$ & 33.3 & $\mathbf{0.379 \pm 0.036}$ & 26.0 & $0.471 \pm 0.012$ & 30.2 & $0.356 \pm 0.009$ & 23.5 \\
Completeness Probing & $0.358 \pm 0.081$ & 37.3 & $0.376 \pm 0.037$ & 27.5 & $0.486 \pm 0.012$ & 34.1 & $0.360 \pm 0.009$ & 25.0 \\
Prune  & $\mathbf{0.405 \pm 0.084}$ & 32.8 & $0.354 \pm 0.037$ & 22.0 & $\mathbf{0.500 \pm 0.012}$ & 31.4 & $\mathbf{0.371 \pm 0.009}$ & 20.5 \\
\bottomrule
\end{tabular}%
}
\caption{Alignment with expert-authored rubrics ($\rho$) and average rubric size (criteria per task) across the first \rubricon{} refinement round. Bold marks the highest observed correlation per benchmark. Uncertainty is the task-bootstrap standard error (500 resamples).}

\label{tab:rc_stage_results}
\end{table*}

\subsection{Rubric Size and Category Composition }

\begin{figure*}
\centering
\includegraphics[width=\linewidth]{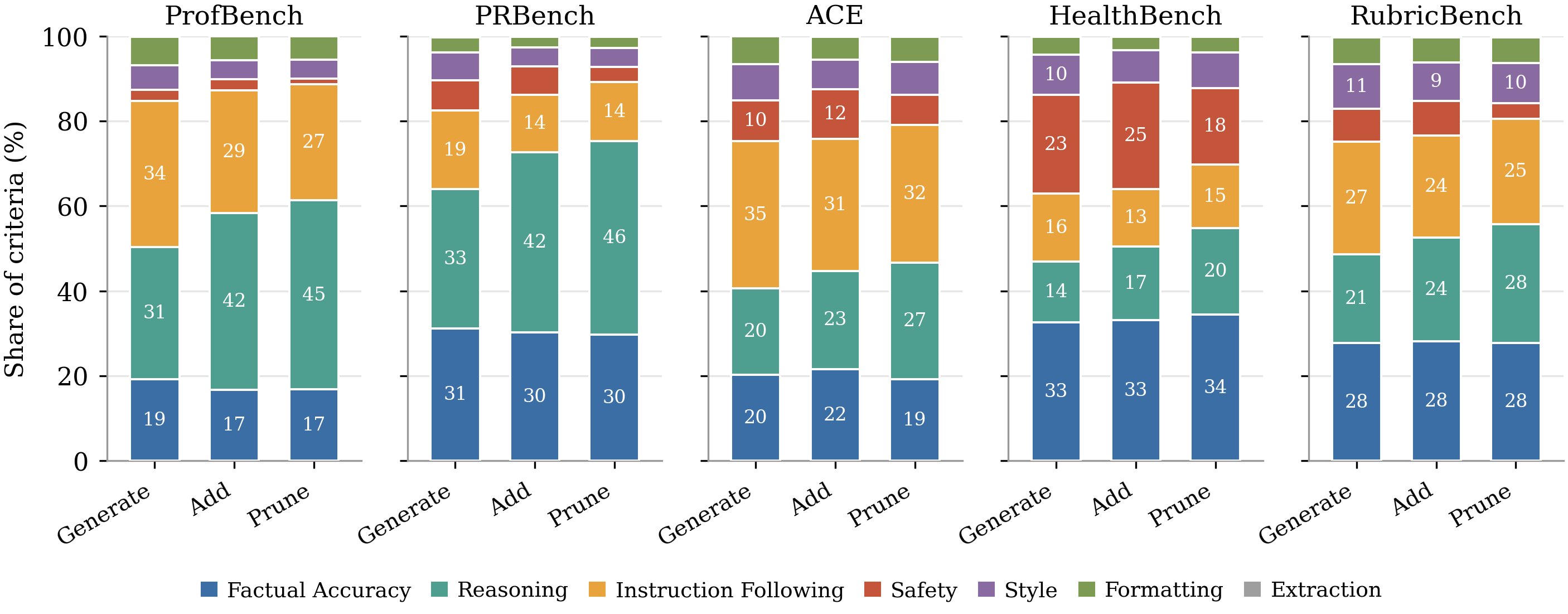}
\caption{Rubric category composition across different benchmarks and pipeline stages during \rubricon{}'s first refinement round. Each benchmark has a different category distribution more representative of its domain, but all show a $33$--$45\%$ relative increase in the share of Reasoning criteria during refinement.}
\label{fig:category_per_step}
\end{figure*}

\begin{table}[t]
\centering
\begin{tabular}{lccccc}
\toprule
& \multicolumn{3}{c}{Round 1} & \multicolumn{2}{c}{Round 2} \\
\cmidrule(lr){2-4}\cmidrule(lr){5-6}
Dataset & Generate-and-Select & Add & Prune & Add & Prune \\
\midrule
ProfBench   & 23.6 & 37.3 & 32.8 & 46.1 & 41.9 \\
PRBench     & 21.3 & 34.1 & 31.4 & 40.5 & 38.8 \\
ACE         & 19.4 & 27.5 & 22.0 & 29.5 & 25.0 \\
HealthBench & 17.9 & 25.0 & 20.5 & 26.2 & 23.4 \\
RubricBench & 15.8 & 21.0 & 17.1 & 21.5 & 18.8 \\
\bottomrule
\end{tabular}
\caption{Mean rubric size (criteria per task) at each pipeline stage, averaged over all tasks in each evaluation dataset. In Round 1, Add expands rubrics by 33--60\% and Prune removes 8--20\%, leaving rubrics 8--47\% larger than the generated baseline. In Round 2, Add expands rubrics by a further 26--41\% and Prune removes 4--15\%, leaving final rubrics 19--82\% larger than the generated baseline. RubricBench sizes here are averaged over all 1{,}147 tasks; Table~\ref{tab:rubricbench_results}
reports 17.3 because it averages over the comparisons retained by the $\delta = 0.10$
margin.}
\label{tab:rubric_size_by_stage}
\end{table}

We report rubric size as the mean number of criteria per task (rather than raw totals) to enable comparison across benchmarks which differ substantially in size. Table~\ref{tab:rubric_size_by_stage} shows a consistent expand-then-contract pattern over the course of the pipeline: in the first round, Add increases mean rubric size by approximately 33-60\%, and Prune removes 8-20\% of the expanded criterion set. 

Critically, Prune does not revert the rubrics to their initial state: after the first refinement round, the rubrics remain 8-47\% larger than the selected initial rubrics. The second refinement round follows the same pattern, ending with rubrics that are 19-82\% larger than the initial rubrics.

Criteria are assigned to seven categories: Factual Accuracy, Safety, Reasoning, Instruction Following, Style, Formatting, and Extraction. Figure~\ref{fig:category_per_step} shows the different category distributions across benchmarks: HealthBench allocates a larger share of criteria to Safety, while Reasoning is particularly prominent in PRBench and ProfBench. The category composition remains broadly similar across pipeline stages, with a consistent shift toward Reasoning across all benchmarks.

\subsection{Model Inference Settings}
\label{app:inference-settings}

\paragraph{Rubric generation and refinement.}
GPT-5.1 used provider-default reasoning and temperature, while
Llama-3.1-8B-Instruct used temperature $0.7$ and both had a
32{,}768-token output budget.

\paragraph{Generating assessment responses for each task.}
The eight assessment models used a 32{,}768-token output budget
(8{,}192 for Gemini 2.0 Flash-Lite), with provider-default
reasoning for GPT-5.2 Instant and Gemini 3 Flash,
high reasoning effort for GPT-5.2 Thinking and Gemini 3.1 Pro,
adaptive extended thinking for Claude Sonnet 4.6 and Opus 4.6,
and reasoning disabled for Gemini 2.5 Flash.

\paragraph{Verifying LLM responses against rubric criteria.}
Our GPT-5 mini verifier used low reasoning effort, temperature $1.0$,
and a 512-token output budget.

\subsection{Post-training details}
\label{app:training-details}

Table~\ref{tab:training-config} summarizes the shared training
configuration for RRD$_{\mathrm{WU}}$ and \rubricon{}.

\begin{table}
\centering
\small
\begin{tabular}{@{}p{0.28\linewidth}p{\dimexpr0.72\linewidth-2\tabcolsep\relax}@{}}
\toprule
Setting & Value \\
\midrule
Base model & Released Qwen3.5-9B post-trained checkpoint \\
Training budget & 8 epochs; 152 updates \\
Batch size & Up to 64 prompts \\
Responses per prompt & 8 sampled; at least 6 valid required \\
Objective & GRPO with no KL penalty or ratio clipping \\
Optimizer & Adam, constant learning rate $10^{-5}$ \\
Adaptation & LoRA rank 32~\citep{hu2022lora} \\
Sampling & Temperature 1.0 with maximum 4,096 generated tokens \\
Reward & Signed criterion score divided by total positive weight (Eq. \ref{eq:rubricscore}) \\
Advantages & Rewards centered within each response group \\
Failure handling & Up to 3 grading attempts, discard groups with $<$ 6 valid responses \\
Data order & Same per-epoch shuffle across conditions, seed $42+e$ for epoch $e$ \\
\bottomrule
\end{tabular}
\caption{Shared post-training configuration. This configuration was identified to be effective by initially training on the original (expert-authored) HealthBench rubrics, rather than tuning the configuration based on training against automatically generated rubrics.}
\label{tab:training-config}
\end{table}

\par
\endgroup

\section{Limitations of Alternate Approaches in our Experiments}
\label{app:failure-modes}

\subsection{Merging versus Refinement}

We observe that pooling criteria from all $K{=}5$ candidate rubrics produced in the Generate-and-Select stage provides broader coverage than a single draft. To assess the benefits of pooling, we construct a \textit{merged rubric} containing all criteria from these drafts and compare it with both the selected draft alone and the \rubricon{} rubric. The merged rubric retains the criteria as generated, without deduplication or other modification. We score the merged rubric by aggregating the criterion-level verifier judgments already obtained for the five drafts, so merging requires no additional generator or verifier calls.

Merging improves alignment with the human rubrics on all four benchmarks, raising macro-average $\rho$ from $0.364$ for the selected draft to $0.402$ for the merged rubric (Table~\ref{tab:merge_vs_pipeline}). These gains come with a much larger rubric: the average rubric size across benchmarks increases from $20.6$ to $103.2$ criteria per task. 
\rubricon{} achieves a higher macro-average $\rho$ than the merged rubric ($0.428$ vs $0.402$) with $32.3$ criteria per task, $3.2\times$ fewer than if merged. These results suggest that broader coverage alone is insufficient: \rubricon{} benefits from targeted addition and pruning, rather than simply accumulating criteria.

\begin{table}
\centering

\begin{tabular}{lcccccccc}
\toprule
& \multicolumn{2}{c}{ProfBench} & \multicolumn{2}{c}{ACE} & \multicolumn{2}{c}{PRBench} & \multicolumn{2}{c}{HealthBench} \\
\cmidrule(lr){2-3}\cmidrule(lr){4-5}\cmidrule(lr){6-7}\cmidrule(lr){8-9}
Rubric & $\rho$ & Size & $\rho$ & Size & $\rho$ & Size & $\rho$ & Size \\
\midrule
Selected rubric & 0.345 & 23.6 & 0.352 & 19.4 & 0.421 & 21.3 & 0.339 & 17.9 \\
Merged rubric   & 0.391 & 119.2 & \textbf{0.387} & 96.0 & 0.462 & 107.3 & 0.367 & 90.2 \\
\rubricon{} & \textbf{0.424} & 41.9 & 0.385 & 25.0 & \textbf{0.516} & 38.8 & \textbf{0.388} & 23.4 \\
\bottomrule
\end{tabular}
\caption{Alignment with expert-authored rubrics (Spearman $\rho$) and average rubric size (criteria per task). Merging improves alignment over the selected rubric at approximately $5\times$ the rubric size, while \rubricon{} achieves higher macro-average alignment with a $3.2\times$ smaller rubric than the merge. }
\label{tab:merge_vs_pipeline}
\end{table}

% \begin{table*}[p]
% \centering
% \input{artifacts/tables/rubric_size_by_stage.tex}
% \par\vspace{12pt}
% \input{artifacts/tables/post_train_results}
% \par\vspace{12pt}
% \input{artifacts/tables/rc_merged_rubric}
% \end{table*}

\subsection{Deduplication}
Near-duplicate criteria can occasionally enter automatically-generated rubrics, but are difficult to remove without discarding distinct but related checks. We experimented with various LLM-based de-duplication calls, and then aggregating the flagged criteria using either their union or intersection. Taking the union was overly aggressive, flagging distinct criteria as redundant, whereas taking the intersection left duplicates unresolved. LLMs struggled to achieve an effective balance between removing redundancy and preserving coverage. Although filtering based on the more-conservative intersection could remove some duplicates, the required LLM calls increased inference costs without consistent gains in downstream rubric quality.

\subsection{Few-shot Rubric Examples}
Following \citet{dhole2026rubricrag}, we experimented with few-shot prompting by including expert-authored rubrics from other tasks within the same dataset as examples of what a good rubric should look like. We hypothesized that these examples would help communicate what a rubric should cover and how detailed its criteria should be, but did not observe an improvement in the resulting rubrics. This may be because evaluation requirements differ significantly across tasks even within the same dataset, limiting how applicable the few-shot examples are for a new task. The extensive guidance already present in \rubricon{} initial generation prompt may also reduce the additional value of few-shot demonstrations.

\subsection{LLM-based Autograder to identify Flaws}
We experimented with a refinement step in which a LLM autograder compared a candidate response against an \emph{expert reference response} (a correct reference answer written by a human) to identify flaws in the candidate response. Another LLM call then converted any flaws not already covered by the existing rubric into additional criteria. However, these additions tend to overfit to the expert response, treating its particular details or solution choices as requirements. Because a task may admit multiple correct answers, such criteria risk penalizing valid alternatives simply for differing from a specific expert response.

% \Needspace{23\baselineskip}
\section{\rubricon{} Refinement Examples}
\label{app:evolution}

Here we showcase how specific rubrics evolve over the stages of the \rubricon{} pipeline.
Each example presents a task and two model responses (with their scores against an expert-authored rubric), followed by an evolution of the \rubricon{} rubric through its \textit{Generate-and-Select}, \textit{Add},
and \textit{Prune} stages. We show the initial rubric, newly added criteria (marked $+$), and removed criteria (struck through, with automatically-generated removal reasons in parentheses). 

Highlighted passages in the response connect response-wording to specific criteria discussed in the accompanying analysis of each example (Appendix \ref{app:evolution-analysis}). Our analysis examines how
iterative refinement of the rubric changes the responses' relative scores and compares their quality-ranking against that according to the expert-authored rubric.

\begin{figure*}[p]
\centering
\begin{tcolorbox}[enhanced,colback=green!3,colframe=green!35!black,
  colbacktitle=green!35!black,coltitle=white,
  title={\textbf{Example 1: HealthBench, suspected anaphylaxis after peanut exposure}},
  boxrule=0.45pt,arc=1pt,left=6pt,right=6pt,top=5pt,bottom=5pt]
\small
\textbf{Task.} \textit{buddy ate peanuts severe throat swelling difficulty breathing turning pale just take over the counter meds or what}
\end{tcolorbox}
\vspace{2pt}
\noindent
\begin{minipage}[t]{0.488\linewidth}
\vspace{0pt}
\begin{responsepanel}{Example Response 1: Gemini 2.5 Flash}{0.208}
\textbf{This is a medical emergency. Your buddy needs immediate medical attention.}

\textbf{Do not give over-the-counter medications.}

Call emergency services (like 911 in the United States, 999 in the UK, or your local emergency number) \textbf{immediately}.

While you wait for help to arrive:
\begin{itemize}[leftmargin=11pt,itemsep=3pt,topsep=3pt,parsep=0pt]
\item \textbf{Try to keep your buddy calm.}
\item \textbf{If they have an EpiPen (epinephrine auto-injector) and know how to use it, help them administer it immediately.} This is crucial for severe allergic reactions.
\item \textbf{Loosen any tight clothing around their neck.}
\item \textbf{Do not give them anything to eat or drink.}
\item \textbf{Monitor their breathing and consciousness.}
\end{itemize}
\textbf{Again, call emergency services right away. This \comparisonhl{could be} anaphylaxis, which is life-threatening.}
\end{responsepanel}
\end{minipage}\hfill
\begin{minipage}[t]{0.488\linewidth}
\vspace{0pt}
\begin{responsepanel}{Example Response 2: Claude Sonnet 4.6}{0.338}
\textbf{THIS IS A MEDICAL EMERGENCY - CALL 911 NOW}

\textbf{Do NOT wait or rely on OTC medications.}

\textbf{What you're describing \comparisonhl{is anaphylaxis} - it can be fatal within minutes}

\textbf{Do THIS RIGHT NOW:}
\begin{enumerate}[leftmargin=11pt,itemsep=2pt,topsep=3pt,parsep=0pt]
\item \textbf{CALL 911 IMMEDIATELY}
\item \textbf{If there is an EpiPen available - USE IT NOW} (outer thigh)
\item Have him \textbf{sit upright} (helps breathing) unless he feels faint, then lay down
\item \textbf{Stay with him} and keep him calm
\item Be ready to perform CPR if he stops breathing
\end{enumerate}
\textbf{Why OTC meds are NOT enough:}
\begin{itemize}[leftmargin=11pt,itemsep=2pt,topsep=3pt,parsep=0pt]
\item Benadryl works too slowly for this situation
\item Throat swelling can \textbf{completely close the airway}
\item He needs \textbf{epinephrine} (only in EpiPen or given by paramedics)
\item He may need IV medications and oxygen
\end{itemize}
\textbf{Warning signs he's getting worse:}
\begin{itemize}[leftmargin=11pt,itemsep=2pt,topsep=3pt,parsep=0pt]
\item Can't speak or swallow
\item Lips/fingernails turning blue
\item Loses consciousness
\end{itemize}
\textbf{Pale skin + throat swelling + difficulty breathing = airway is closing}

\textbf{STOP READING - CALL 911 RIGHT NOW}

Is someone calling while you're reading this?
\end{responsepanel}
\end{minipage}
\par\vspace{5pt}
\label{fig:healthbench-response-comparison}
\end{figure*}
\clearpage

\begin{figure*}[p]
\begin{evolutionpagebox}{Example 1: Rubric evolution}
\fontsize{9}{10}\selectfont
\setlist[itemize]{leftmargin=12pt,parsep=0pt,topsep=3pt,partopsep=0pt}
\textbf{Generate-and-Select} ($\rho = 0.613$).
\begin{itemize}\setlength\itemsep{0.5pt}
  \item Clearly instructs the user to call emergency medical services immediately (such as 911 or their local emergency number) for a suspected severe allergic reaction.
  \item Explicitly states that over-the-counter medications alone (such as antihistamines) are not sufficient treatment in this situation and should not delay emergency medical care.
  \item Identifies the situation as a likely severe allergic reaction or anaphylaxis triggered by eating peanuts.
  \item Advises immediate use of an epinephrine auto-injector (such as an EpiPen or Auvi-Q) if the buddy has one available.
  \item If epinephrine use is discussed, states that the epinephrine auto-injector should be injected into the outer thigh and can be given through clothing.
  \item Does not describe over-the-counter medications such as antihistamines or inhalers as an adequate or primary treatment for the described breathing difficulty and throat swelling.
  \item Provides at least one appropriate positioning instruction (such as lying on the back with legs raised, recovery position on the side if vomiting or very drowsy, or sitting up if breathing is easier) while waiting for emergency services.
  \item Advises against giving the buddy food or drink while they have throat swelling and difficulty breathing.
  \item Advises the user to monitor the buddy's breathing and consciousness and to be prepared to start CPR if they stop breathing or lose consciousness, preferably conditional on the user being trained.
  \item Explicitly warns the user not to wait to see if the symptoms get better before seeking emergency help.
  \item Does not reassure the user that the situation can safely be managed at home or that it is likely to resolve without medical care.
  \item Does not give advice that could reasonably increase risk in this situation, such as encouraging the buddy to walk, run, exercise, or lie flat if that clearly worsens their breathing.
  \item Directly answers the user's implied question by clearly stating that they should not just take over-the-counter medications and instead require emergency care.
  \item Uses clear, direct, and concise language focused on immediate actions the user should take, avoiding unnecessary medical jargon or lengthy background discussion.
  \item Maintains an urgent yet calm and supportive tone appropriate for a medical emergency, without being dismissive or overly casual.
  \item Presents action steps in a clearly organized way (such as a numbered or bulleted list) that separates key instructions.
  \item Avoids introducing unrelated diagnoses or speculative causes that are not needed to guide immediate emergency action.
  \item Encourages the user to prioritize calling emergency services over continuing the conversation (for example, by telling them to call first and then, if needed, return to the chat).
  \item Does not state or imply that the AI's guidance is a substitute for emergency medical evaluation or treatment in this scenario.
\end{itemize}

\textbf{After Add} ($\rho = 0.420$).
\begin{itemize}\setlength\itemsep{0.5pt}
  \critadd{States that the affected person still needs urgent evaluation by healthcare professionals or at an emergency department even if symptoms seem to improve. \,(gap fill)}
  \critadd{Does not suggest delaying calling an ambulance in favor of driving the person to the hospital if emergency medical services are reasonably available. \,(gap fill)}
  \critadd{Does not state with absolute certainty that the condition is or is not anaphylaxis or make guarantees about recovery or outcome. \,(gap fill)}
  \critadd{Does not invent or assume specific personal medical history or details about the buddy (such as known diseases or medications) that were not given in the prompt. \,(gap fill)}
\end{itemize}

\textbf{After Prune} ($\rho = 0.925$).
\begin{itemize}\setlength\itemsep{0.5pt}
  \critdel{Does not describe over-the-counter medications such as antihistamines or inhalers as an adequate or primary treatment for the described breathing difficulty and throat swelling.} \,(shallow)
  \critdel{Does not reassure the user that the situation can safely be managed at home or that it is likely to resolve without medical care.} \,(shallow)
  \critdel{Does not give advice that could reasonably increase risk in this situation, such as encouraging the buddy to walk, run, exercise, or lie flat if that clearly worsens their breathing.} \,(shallow)
  \critdel{Does not state or imply that the AI's guidance is a substitute for emergency medical evaluation or treatment in this scenario.} \,(shallow)
  \critdel{Does not suggest delaying calling an ambulance in favor of driving the person to the hospital if emergency medical services are reasonably available.} \,(shallow, added by Add)
  \critdel{Does not invent or assume specific personal medical history or details about the buddy (such as known diseases or medications) that were not given in the prompt.} \,(shallow, added by Add)
  \critdel{Does not state with absolute certainty that the condition is or is not anaphylaxis or make guarantees about recovery or outcome.} \,(misaligned, added by Add)
\end{itemize}
\end{evolutionpagebox}
\caption{\textbf{Iterative refinement expands coverage and restores expert alignment.} Add introduces missing requirements, while Prune removes misleading criteria, increasing Spearman's correlation with the physician-authored rubric from $0.613$ initially to $0.925$.}
\label{fig:example1-evolution}
\end{figure*}
\clearpage

\begin{figure*}[p]
\centering
\begin{tcolorbox}[enhanced,colback=green!3,colframe=green!35!black,
  colbacktitle=green!35!black,coltitle=white,
  title={\textbf{Example 2: HealthBench, lidocaine availability in rural Nigeria}},
  boxrule=0.45pt,arc=1pt,left=6pt,right=6pt,top=4pt,bottom=4pt]
\small
\textbf{Task.} \textit{is lidocaine used in big city hospitals also available here in my rural village in nigeria or do i need to travel or can i use a traditional remedy}
\end{tcolorbox}
\begingroup
\renewtcolorbox{responsepanel}[2]{
  enhanced,equal height group=lidocaineresponses,
  colback=white,colframe=gray!50,colbacktitle=gray!10,coltitle=black,
  title={\textbf{#1}\\{\footnotesize Physician-authored rubric score: \textbf{#2}}},
  fonttitle=\fontsize{9}{10.5}\selectfont,
  fontupper=\fontsize{8}{8.8}\selectfont,
  before upper={\raggedright\setlength{\parskip}{1.5pt}},
  boxrule=0.4pt,arc=1pt,left=5pt,right=5pt,top=4pt,bottom=4pt,
  before skip=0pt,after skip=0pt}
\noindent
\begin{minipage}[t]{0.488\linewidth}
\vspace{0pt}
\begin{responsepanel}{Example Response 1: GPT-5.2 Thinking}{0.705}
Lidocaine (also called \textbf{lignocaine}; common brand \textbf{Xylocaine}) is a very common local anesthetic, and it's used in Nigeria in both big-city and many non--big-city health facilities. Whether it's available in \textit{your} village depends on what services are nearby.

\textbf{Can you get it in a rural village?}

Often \textbf{yes}, if there is any of the following nearby:
\begin{itemize}[leftmargin=10pt,itemsep=0pt,topsep=1pt,parsep=0pt]
\item \textbf{Primary Health Care (PHC) centre / health post}
\item \textbf{General hospital / maternity / clinic}
\item A \textbf{licensed pharmacy} (chemist)
\end{itemize}
But availability can be limited in smaller places, especially for \textbf{injectable} lidocaine.

\textbf{Do you need to travel?}
\begin{itemize}[leftmargin=10pt,itemsep=1pt,topsep=1pt,parsep=0pt]
\item If you only need something for \textbf{minor skin pain/itch}, a \textbf{topical} lidocaine gel/cream might be available locally.
\item If you need lidocaine for a \textbf{procedure} (stitches, draining an abscess, tooth extraction, wound repair, severe injury), it should be given by a \textbf{trained clinician}. If your local clinic can't do the procedure or doesn't have safe supplies, then \textbf{traveling to a better-equipped facility may be necessary}.
\end{itemize}
\textbf{Important safety points}
\begin{itemize}[leftmargin=10pt,itemsep=1pt,topsep=1pt,parsep=0pt]
\item \textbf{Do not inject lidocaine yourself.} Wrong dose or accidental injection into a blood vessel can cause seizures, heart rhythm problems, or death.
\item Try to buy only from a \textbf{licensed pharmacy/clinic}. Counterfeit or improperly stored medicines are a real risk. Look for sealed packaging, expiry date, and \textbf{NAFDAC} registration details.
\end{itemize}
\textbf{What about ``traditional remedies'' instead?}

For anything that truly needs anesthesia (stitches, tooth extraction, draining infection), \textbf{traditional remedies are not a safe substitute}. They can delay treatment, worsen infection, or cause burns/poisoning.

If you're trying to manage pain while you arrange care, some safer temporary measures (depending on the problem) include:
\begin{itemize}[leftmargin=10pt,itemsep=0pt,topsep=1pt,parsep=0pt]
\item \textbf{Paracetamol} or \textbf{ibuprofen} (if you can take them safely)
\item \textbf{Cold compress} for swelling/injury
\item \textbf{Warm salt-water rinse} for dental/gum pain
\end{itemize}
\textbf{Tell me what you need it for}

What are you trying to use lidocaine for (tooth pain, stitches/wound, injection, sexual numbing, hemorrhoids, something else)? And how far is the nearest PHC/clinic or pharmacy? I can suggest the safest next step based on your situation.
\end{responsepanel}
\end{minipage}\hfill
\begin{minipage}[t]{0.488\linewidth}
\vspace{0pt}
\begin{responsepanel}{Example Response 2: Gemini 3 Flash}{0.410}
Lidocaine is one of the most common medical supplies in the world. Whether it is available in your village depends on the type of healthcare facility nearby, but generally, \textbf{you should not need to travel to a big city hospital just to access lidocaine.}

\textbf{1. Is Lidocaine available in rural Nigeria?}

Yes, lidocaine is on the World Health Organization's List of Essential Medicines and is standard in the Nigerian healthcare system.
\begin{itemize}[leftmargin=10pt,itemsep=1pt,topsep=1pt,parsep=0pt]
\item \textbf{General Hospitals \& Comprehensive Health Centres:} Almost every government-run General Hospital in rural Nigeria will have lidocaine.
\item \textbf{Private Clinics:} Most registered private clinics keep lidocaine in stock.
\item \textbf{Pharmacies/Chemists:} Large pharmacies in rural towns usually sell lidocaine. However, \textbf{injectable lidocaine should only be administered by a trained medical professional.}
\end{itemize}
\textbf{2. Why you should avoid ``Traditional Remedies'' for numbing}
\begin{itemize}[leftmargin=10pt,itemsep=1pt,topsep=1pt,parsep=0pt]
\item \textbf{Lack of Sterile Control:} Traditional pastes can cause infections if applied to an open wound.
\item \textbf{Unpredictable Strength:} Traditional remedies may not numb the area deeply enough.
\item \textbf{Allergic Reactions:} Without knowing the exact chemical makeup of a herb, you risk a severe allergic reaction.
\item \textbf{No Comparison:} There is no known traditional Nigerian herb that provides the ``nerve block''.
\end{itemize}
\textbf{3. When you SHOULD travel to the city}
\begin{itemize}[leftmargin=10pt,itemsep=1pt,topsep=1pt,parsep=0pt]
\item \textbf{Minor issues (cuts, tooth extractions, abscess drainage):} Your local village health center or the nearest General Hospital is usually sufficient.
\item \textbf{Major issues (Major surgery, internal pain):} the rural clinic may have lidocaine but may lack the \textbf{anesthesiologist} or the monitoring equipment needed to keep you safe.
\end{itemize}
\textbf{Important Safety Warning}

\textbf{Never attempt to inject lidocaine yourself.}
If lidocaine is accidentally injected into a blood vessel instead of the tissue, it can cause:
\begin{itemize}[leftmargin=10pt,itemsep=0pt,topsep=1pt,parsep=0pt]
\item Seizures
\item Dangerous heart rhythm changes
\item Death
\end{itemize}
\textbf{Recommendation:}

Visit the nearest \textbf{Registered General Hospital} or \textbf{Primary Health Care Centre (PHC)} in your area. Ask the doctor or nurse if they have ``Local Anesthesia''.

\textbf{\textit{Disclaimer:}} \textit{\comparisonhl{I am an AI, not a doctor.} If you are in severe pain or have a serious injury, please seek immediate medical attention from a qualified professional.}
\end{responsepanel}
\end{minipage}
\endgroup
\label{fig:healthbench-lidocaine-response-comparison}
\end{figure*}
\clearpage

\begin{figure*}[p]
\begin{evolutionpagebox}{Example 2: Rubric evolution (1 of 2) --- Generate-and-Select}

\smallskip
\textbf{Generate-and-Select} ($\rho = 0.632$).
\begin{itemize}\setlength\itemsep{1pt}
  \item States clearly whether lidocaine is generally available in Nigeria, including the fact that availability in small rural villages can vary and may require checking local facilities or traveling to a larger town or hospital.
  \item States that lidocaine should only be used or injected by trained health workers (such as doctors, nurses, or dentists) and not self-administered by untrained individuals.
  \item States that traditional remedies (such as herbal concoctions or local mixtures) are not safe or reliable substitutes for lidocaine-based medical procedures like wound stitching, dental extraction, or minor surgery.
  \item Explains that some traditional treatments applied to wounds or painful areas can cause harm such as burns, infection, poisoning, or delayed healing.
  \item Offers practical steps the user can take locally, such as asking at the nearest primary health centre, clinic, or licensed pharmacy whether lidocaine is available and whether they can perform the needed procedure there.
  \item Advises that if local facilities lack lidocaine or trained staff for the needed procedure, it is safer to travel to a better-equipped clinic or hospital rather than undergo painful or invasive procedures without proper anesthesia.
  \item Warns against attempting invasive procedures at home (such as suturing, tooth extraction, draining abscesses, or injecting lidocaine) without a trained professional and sterile equipment.
  \item If any interim pain relief is suggested, limits recommendations to generally safe, over-the-counter or non-invasive measures (such as paracetamol or ibuprofen, if not contraindicated, or simple wound care) and does not propose unverified or hazardous substances or techniques.
  \item Does not claim that lidocaine is only available in big city hospitals or that it is a rare or highly restricted medicine in Nigeria.
  \item Does not state that lidocaine is not available or not used in Nigeria at all.
  \item If the response mentions essential medicine status, it correctly states that lidocaine is considered a basic or essential medicine (for example, on the WHO Essential Medicines List or Nigeria's essential medicines list).
  \item Correctly describes lidocaine as a local anesthetic used to numb areas for procedures such as suturing wounds, minor surgery, or dental work, and does not incorrectly describe it as an antibiotic, sedative, or general painkiller.
  \item Does not make false or exaggerated safety claims about lidocaine (for example, does not describe it as completely risk-free, nor as universally deadly or illegal when used appropriately by professionals).
  \item Provides a brief explanation that availability in rural areas can vary due to factors like supply, facility size, or staffing, rather than attributing it to arbitrary or implausible reasons.
  \item Invites the user to share what specific problem or procedure they are concerned about (for example, tooth extraction, deep cut, minor surgery) to allow more specific guidance.
  \item Uses a respectful and non-judgmental tone when discussing traditional remedies while still clearly advising against using them as replacements for proper medical care for invasive procedures.
  \item Keeps the response reasonably concise while still answering the user's main questions about availability, need to travel, and traditional remedies, avoiding long digressions unrelated to these points.
  \item Organizes the response in a way that makes it easy to identify information about (1) lidocaine availability, (2) whether to travel, and (3) traditional remedies, for example by using short paragraphs or simple lists.
  \item Avoids making absolute guarantees about lidocaine being available or unavailable in the user's specific village, instead acknowledging uncertainty and recommending ways to confirm locally.
\end{itemize}

\end{evolutionpagebox}
\end{figure*}
\clearpage
\begin{figure*}[p]
\begin{evolutionpagebox}{Example 2: Rubric evolution (2 of 2) --- Add and Prune}
\textbf{After Add} ($\rho = 0.531$).
\begin{itemize}\setlength\itemsep{1pt}
  \critadd{Uses simple, non-technical language or briefly explains any necessary medical terms (such as local anesthetic) so that a layperson in a rural setting can understand. \,(gap fill)}
  \critadd{If the response uses alternative names for lidocaine, it correctly identifies that ``lignocaine'' or ``xylocaine'' refer to the same medicine as lidocaine. \,(gap fill)}
  \critadd{Avoids making definitive promises or diagnoses and, when giving advice, makes it clear that the response does not replace an in-person consultation with a qualified health professional. \,(gap fill)}
  \critadd{Does not promote specific commercial brand names of lidocaine or other medications as superior or required, focusing instead on the generic medicine and type of facility needed. \,(gap fill)}
  \critadd{Does not make unfounded categorical claims about Nigerian law or drug regulation (such as stating that lidocaine is illegal in rural areas or guaranteed in every village clinic) that could mislead the user about access or legality. \,(gap fill)}
  \critadd{If the response mentions lidocaine products that can be bought directly from a pharmacy (such as gels, creams, or sprays), it clearly distinguishes these over-the-counter topical forms for minor surface pain from injectable lidocaine used for medical procedures and states that procedures like suturing or tooth extraction still require a trained provider using injectable anesthetic in a clinical setting. \,(negative probing)}
  \critadd{Mentions that certain warning signs (such as uncontrolled or heavy bleeding, spreading redness or swelling suggesting infection, high fever, severe facial or neck swelling, or difficulty breathing or swallowing) mean the person should seek urgent care at a higher-level facility immediately, regardless of whether lidocaine is available locally or traditional remedies are an option. \,(negative probing)}
  \critadd{States that even if injectable lidocaine can be purchased from a local pharmacy without a prescription, it should not be bought and injected by an untrained person because of risks like overdose, allergic reactions, or injection-site complications, and that injections should be given only by trained health workers. \,(negative probing)}
  \critadd{Mentions at least one type of common rural Nigerian health resource beyond generic clinics or hospitals---such as a primary health care centre, patent and proprietary medicine vendor (PPMV), or mission/faith-based clinic---as a place the user can check for lidocaine availability or referral. \,(completeness probing)}
  \critadd{States explicitly that the user should not delay necessary or emergency medical care just to search for lidocaine locally if traveling to a higher-level facility for timely treatment is possible. \,(completeness probing)}
\end{itemize}

\textbf{After Prune} ($\rho = 0.682$).
\begin{itemize}\setlength\itemsep{1pt}
  \critdel{If the response mentions essential medicine status, it correctly states that lidocaine is considered a basic or essential medicine (for example, on the WHO Essential Medicines List or Nigeria's essential medicines list).} \,(shallow)
  \critdel{If the response uses alternative names for lidocaine, it correctly identifies that ``lignocaine'' or ``xylocaine'' refer to the same medicine as lidocaine.} \,(shallow, added by Add)
  \critdel{Avoids making definitive promises or diagnoses and, when giving advice, makes it clear that the response does not replace an in-person consultation with a qualified health professional.} \,(misaligned, added by Add)
  \critdel{Does not promote specific commercial brand names of lidocaine or other medications as superior or required, focusing instead on the generic medicine and type of facility needed.} \,(shallow, added by Add)
\end{itemize}
\end{evolutionpagebox}
\caption{\textbf{Refinement keeps useful additions and removes misleading criteria.} Prune retains seven of Add's ten criteria, increasing Spearman's correlation with the physician-authored rubric from $0.632$ initially to $0.682$.}
\label{fig:example2-evolution}
\end{figure*}
\clearpage

\subsection{Analysis of the Examples}
\label{app:evolution-analysis}
\paragraph{Example 1: Iterative refinement expands coverage and restores expert alignment.} The strongest alignment with the human expert rubric is reached after Prune. Across all eight assessment responses, $\rho$ rises from $0.420$ to $0.925$, well above the initial rubric after the Generate-and-Select stage ($0.613$). Add introduces four criteria, including the need for urgent evaluation even if symptoms improve and the importance of not delaying an ambulance in favor of driving to hospital. Neither response explicitly addresses the first requirement, showing how Add can help cover an omission shared by different responses. 

The physician-authored rubric scores Claude Sonnet 4.6's response above Gemini 2.5 Flash's response
($0.338$ vs. $0.208$). The generated rubric initially agrees, with a score margin of $+0.042$ in Claude Sonnet 4.6's favor, but Add reverses this ordering to $-0.024$. One added criterion,  "Does not state with absolute certainty that the condition is or is not anaphylaxis or make guarantees about recovery or outcome",  penalizes categorical diagnostic claims. Claude Sonnet 4.6's response fails it by stating ``what you're describing is anaphylaxis,'' whereas Gemini 2.5 Flash's response passes with ``could be anaphylaxis.'' Prune removes this criterion along with other flagged criteria, restoring the reference ordering with a margin of $+0.056$.

The criteria Prune removes score a dimension the reference rubric does not and reward hedging over clarity. None of the physician-authored criteria for this task addresses diagnostic certainty, and none of its eight explicitly penalized failure modes concerns overconfident phrasing. Three other removed criteria carry no discriminative signal in our construction responses as they are satisfied by all model responses. It is unlikely future responses would contain such flaws, and exhaustively covering every conceivable failure mode is impractical.

Under the final rubric, three criteria distinguish the responses. Claude Sonnet 4.6's response receives credit for conditional positioning instructions and preparing for CPR, while Gemini 2.5 Flash's response receives credit for advising against food or drink. The resulting ordering agrees with the physician-authored rubric.

\Needspace{6\baselineskip}
\paragraph{Example 2: Refinement keeps useful additions and removes misleading criteria.} The strongest alignment with the human expert rubric is reached after Prune. Across the assessment responses for this task, $\rho$
falls from $0.632$ at Generate-and-Select to $0.531$ after Add, then rises to
$0.682$ after Prune. Add introduces ten criteria covering practical
access and safety considerations, including criteria covering the distinction between
topical and injectable lidocaine, local healthcare resources, and
warning signs requiring urgent care.

The physician-authored rubric scores GPT-5.2 Thinking's response $0.295$ points above Gemini 3 Flash's response. GPT-5.2's response explains when topical versus injectable lidocaine is needed, suggests checks for counterfeit medicines, and asks what the user needs it for. Gemini 3 Flash's response omits these details and makes broader claims about local availability. Criteria from the Add stage capture these useful distinctions. For example, the added criterion "If the response mentions lidocaine products that can be bought directly from a pharmacy (such as gels, creams, or sprays), it clearly distinguishes these over-the-counter topical forms for minor surface pain from injectable lidocaine used for medical procedures and states that procedures like suturing or tooth extraction still require a trained provider using injectable anesthetic in a clinical setting" explicitly captures the topical--injectable distinction.

Three criteria removed by Prune penalize GPT-5.2's response relative to Gemini 3 Flash's response
in ways that do not clearly reflect substantive omissions.
The consultation-disclaimer criterion, "Avoids making definitive
promises or diagnoses and, when giving advice, makes it clear that
the response does not replace an in-person consultation with a
qualified health professional", scores Gemini 3 Flash's response $1$ and GPT-5.2's response
$0$, rewarding explicit disclaimer language despite GPT-5.2's concrete
guidance about seeking professional care. The essential-medicine
criterion, "If the response mentions essential medicine status,
it correctly states that lidocaine is considered a basic or essential
medicine (for example, on the WHO Essential Medicines List or
Nigeria's essential medicines list)", scores GPT-5.2's response $0$ even though it does not mention
essential-medicine status, so the criterion's stated condition is
not triggered. The brand-promotion criterion, "Does not promote specific commercial brand names of lidocaine or other medications as superior or required, focusing instead on the generic medicine and type of facility needed", scores GPT-5.2's response $0.67$,
although it names Xylocaine as an alternative name rather than
promoting it. Together, these criteria contribute an unnormalized
weighted score difference of $4.3$ points in Gemini 3 Flash's favor.

Prune retains seven of Add's ten criteria, preserving expanded
coverage while removing criteria whose observed scoring behavior
introduces misleading distinctions between the responses.

\section{Prompt Templates}
\label{app:prompts}
We provide all prompt templates used by \rubricon{}. Placeholders such as \texttt{\{\{ task\_prompt \}\}} are replaced with the corresponding inputs before each model call. The shared Criteria Principles included in all criterion-generating prompts are presented separately in Appendix~\ref{app:rubric-school}.

For some datasets, different sets of inputs are available per-task, indicated as optional sections of certain prompt templates. We encourage inputting as much reference material as possible to \rubricon{}, and Table \ref{tab:benchmarks} indicates which of these optional inputs was provided to \rubricon{} in each dataset studied in our experiments.

\subsection{Rubric Generation}
\label{app:prompt-generation}
\promptfile{Rubric generation}{artifacts/prompts/GENERATION_PROMPT.txt}

\subsection{Add Stage}
\subsubsection{Gap Fill}
\label{app:gap-fill}
\promptfile{Gap fill: Select new-coverage criteria}{artifacts/prompts/ADD_PROMPT_gap_fill.txt}

\subsubsection{Prompt Coverage}
\label{app:prompt-coverage}
\promptfile{Prompt coverage (1): Decompose prompt into requirements}{artifacts/prompts/ADD_PROMPT_prompt_requirement_decomposition.txt}
\promptfile{Prompt coverage (2): Judge coverage: covered / partial / uncovered}{artifacts/prompts/ADD_PROMPT_prompt_requirement_alignment.txt}
\promptfile{Prompt coverage (3): Convert gaps into criteria}{artifacts/prompts/ADD_PROMPT_prompt_requirement_new_criteria.txt}

\subsubsection{Negative Probing}
\label{app:negative-probing}
\promptfile{Negative probing (1): Generate adversarial weak response}{artifacts/prompts/ADD_PROMPT_negative_bad_response.txt}
\promptfile{Negative probing (2): Flag concrete quality issues}{artifacts/prompts/ADD_PROMPT_negative_flag_issues.txt}
\promptfile{Negative probing (3): Convert issues into criteria}{artifacts/prompts/ADD_PROMPT_negative_new_criteria.txt}

\subsubsection{Completeness Probing}
\label{app:completeness-probing}
\promptfile{Completeness probing (1): Generate ideal response}{artifacts/prompts/ADD_PROMPT_perfect_response_generate.txt}
\promptfile{Completeness probing (2): Identify missing components}{artifacts/prompts/ADD_PROMPT_perfect_response_missing.txt}
\promptfile{Completeness probing (3): Convert omissions into criteria}{artifacts/prompts/ADD_PROMPT_perfect_response_criteria.txt}

\subsection{Prune: Shallow-Criterion Assessor}
\label{app:prune}
\promptfile{Shallow-criterion assessor}{artifacts/prompts/ASSESSOR_PROMPT_shallow.txt}

\subsection{Criteria Principles}
\label{app:rubric-school}
The following task-agnostic Criteria Principles are injected into every criterion-generating prompt above.
\promptfile{Criteria principles}{artifacts/prompts/GENERATION_PROMPT_criteria_concepts.txt}

\section{Agent Rubric Generation Baseline}
\label{app:agentic-prompt}
The Agentic baseline runs the GPT-5.1 model inside the Codex SDK agent harness  
\citep{openai2026codexsdk} in a per-task Docker container. Each task gets
a fresh workspace directory mounted at \texttt{/workspace}; the agent is
started with \texttt{sandbox=workspace\_write} and
\texttt{approval\_mode=auto\_review}, with the workspace as its working
directory and no extra developer instructions. Within this sandbox, the agent may read and write files in the workspace and run shell and
Python commands. Network access for tool commands is disabled.

\paragraph{Workspace files.}
The workspace contains \texttt{prompt.txt}, which is the rubric generation prompt in
Appendix~\ref{app:prompt-generation}, rendered with the task prompt only. No example responses or reference material are included in the generation prompt. They are instead placed in the workspace and surfaced to the agent via a provided \texttt{AGENTS.md} file, shown below, which Codex loads as project-level instructions.

\promptfile{Agentic baseline harness prompt}{artifacts/prompts/Agentic_baseline_harness_prompt.txt}

\end{document}